\documentclass[letterpaper]{article} 
\usepackage[]{aaai2026}  
\usepackage{times}  
\usepackage{helvet}  
\usepackage{courier}  
\usepackage[hyphens]{url}  
\usepackage{graphicx} 
\usepackage{natbib, multirow}  
\usepackage{caption} 
\usepackage[ruled, lined]{algorithm2e}
\usepackage{subfigure}
\usepackage{amsfonts, bm, amsmath}
\usepackage{amsmath}
\usepackage{subcaption, array, tabularx}
\usepackage{amssymb, bm}
\usepackage{pifont}
\usepackage{ragged2e}
\newcommand{\cmark}{\ding{51}}%
\newcommand{\xmark}{\ding{55}}%
\usepackage{adjustbox}
\usepackage{makecell}
\usepackage{scrextend}
\usepackage{svg}

\title{CART: Compositional AutoRegressive Transformer for Image Generation}
\author {
    Siddharth Roheda,
    Rohit Chowdhury,
    Aniruddha Bala,
    Rohan Jaiswal
}
\affiliations {
    Samsung Research Institute\\
    Bangalore, India\\
    \{sid.roheda, rohit.c, aniruddha.b, r.jaiswal\}@samsung.com
}

\usepackage{bibentry}
\usepackage{pdfpages}
\usepackage{graphicx, multicol}

\begin{document}
\twocolumn[{%
\renewcommand\twocolumn[1][]{#1}%
\maketitle
\begin{center}
   \centering
   \captionsetup{type=figure}
   \includegraphics[width=.8\textwidth]{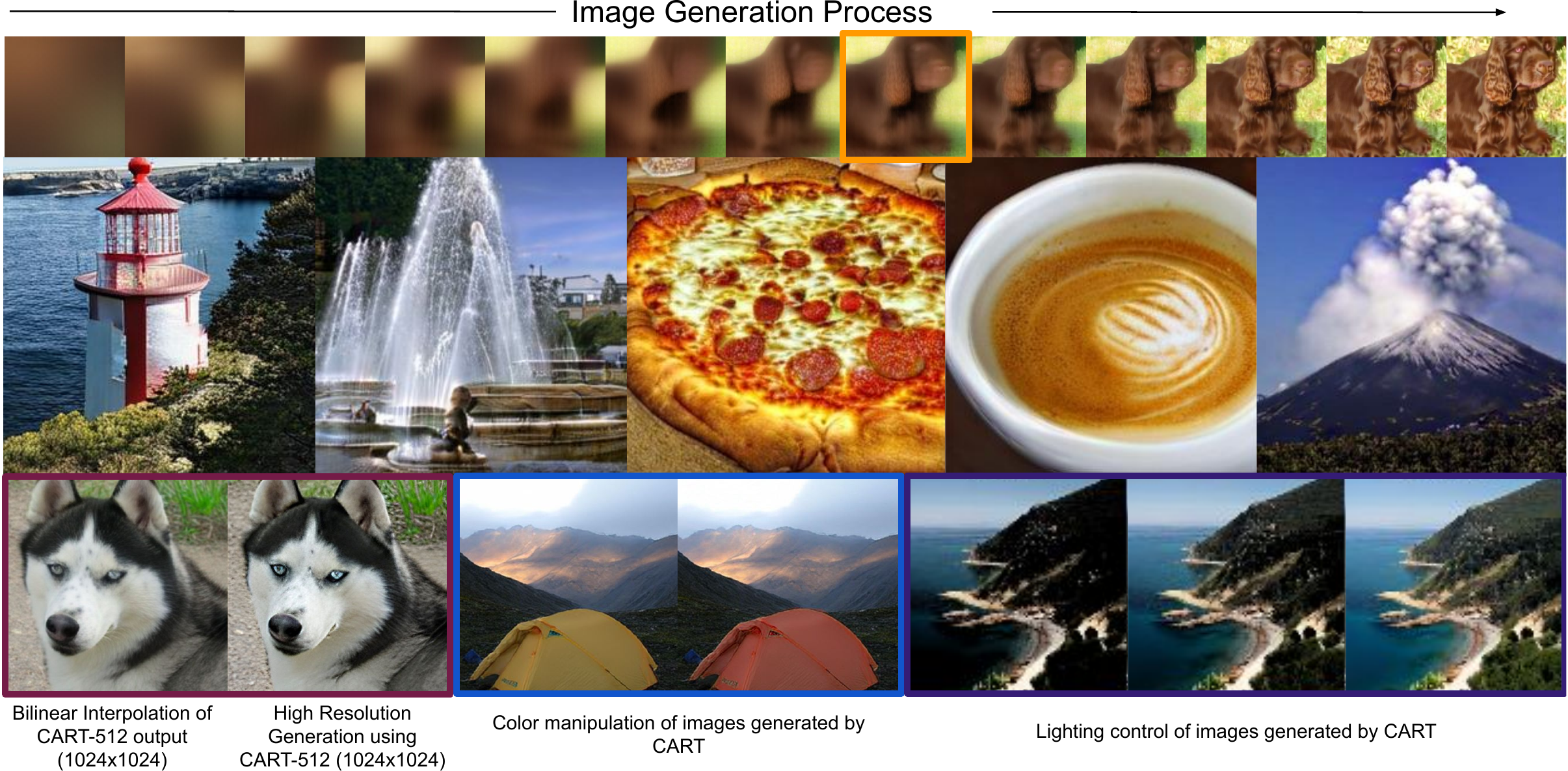}
   \captionof{figure}{Top row: Image Generation process using CART. Orange box marks generated base component. Middle Row: Generated image samples using CART. Bottom row: Applications of CART: High resolution generation without re-training (Magenta), Recoloring objects in scenes (Blue), Lighting control (Purple)}
   \label{fig:teaser}
\end{center}%
}]



\begin{abstract}
We propose a novel Auto-Regressive (AR) image generation approach that models images as hierarchical compositions of interpretable visual layers. While AR models have achieved transformative success in language modeling, replicating this success in vision tasks remains challenging due to inherent spatial dependencies in images. Addressing the unique challenges of vision tasks, our method (CART) adds image details iteratively via semantically meaningful decompositions. We demonstrate the flexibility and generality of CART by applying it across three distinct decomposition strategies: (i) Base-Detail Decomposition (Mumford-Shah smoothness), (ii) Intrinsic Decomposition (albedo/shading), and (iii) Specularity Decomposition (diffuse/specular). This “next-detail" strategy outperforms traditional “next-token" and “next-scale" approaches, improving controllability, semantic interpretability, and resolution scalability. Experiments show CART generates visually compelling results while enabling structured image manipulation, opening new directions for controllable generative modeling via physically or perceptually motivated image factorization.

\end{abstract}    
\section{Introduction}
\label{sec:intro}
Recent advancements in Generative AI for image synthesis have garnered significant interest across research and industry. Conventional approaches including Generative Adversarial Networks (GANs) \cite{goodfellow2020generative, mirza2014conditional} and Variational Auto Encoders (VAEs) \cite{kingma2013auto, shao2020controlvae} typically generate entire scenes in a single pass. Recent research has introduced step-wise approaches where each step incorporates a subset of details. Diffusion-based methods \cite{ho2020denoising, song2020denoising} initiate with noise and employ denoising models to progressively reveal coherent images. Similarly, Auto-Regressive (AR) models \cite{van2016pixel, salimans2017pixelcnn++, gregor2015draw, parmar2018image} tackle generation in a patch-wise manner. Image generation models like VQGAN \cite{esser2021taming} and DALLE \cite{ramesh2021zero} aim to parallel the success of AR models in Large Language Modelling (LLMs) by using visual tokenizers that convert images into 2D token grids enabling next-token prediction.

Despite success of AR models in Natural Language Processing (NLP), achieving similar vision advancements remains challenging. Recent studies in AR \cite{tian2024visual} highlight that token prediction sequence can significantly impact performance. VAR \cite{tian2024visual} adopts a multi-scale tokenization approach, where token maps at different scales are created within the encoded latent space. A transformer is then trained to predict the next higher-resolution token map, while conditioning on the previously generated token maps. This “next-scale" strategy enables progressive resolution expansion, improving upon raster-scan tokenization. However, while VAR enhances scalability, it refines both global structures and fine details simultaneously at each scale, without explicitly disentangling them. This entanglement of structural and textural features limits fine-grained control over generated image characteristics. Furthermore, such an intertwined representation necessitates retraining or fine-tuning whenever the target generation resolution deviates from that used during VAR training, limiting its flexibility across resolutions.

To address these limitations, we draw inspiration from human perception and visual content creation, which fundamentally follow a compositional approach. For instance, artists typically begin by outlining spatial layouts and global structures, then progressively refine color, textures, and details. Motivated by this process, we propose a novel compositional AR framework that synthesizes images by sequentially predicting constituent visual factors, ranging from structural layouts to appearance refinements.
Our approach decomposes training images into physically relevant “base" and “detail" components, encoding them into multi-scale detail token-maps. AR processing initiates with 1×1 tokens, predicting successive token-maps to construct a base component at the target resolution. The model then predicts detail components, incrementally layering them to enhance the base image. Such a “next-detail” generation approach enables the model to synthesize images with significantly higher detail compared to state-of-the-art methods such as \cite{tian2024visual} and \cite{sun2024autoregressive}. In addition, it supports training-free high-resolution image generation and facilitates super-resolution of low-quality inputs.

The \textbf{Contributions} of this paper include: 
\begin{itemize}
    \item A novel \textbf{iterative image generation} approach aligning with natural image formation order.
    \item A \textbf{hierarchical tokenization strategy} to quantize an image into base and detail layers.
    \item \textbf{High-resolution generation and Super-Resolution without retraining}, demonstrating versatility.
    \item \textbf{Fine-grained control} over image characteristics such as textures, colors, and lighting.
\end{itemize}

\section{Related Work}
\label{sec:rel_wrk}

\subsection{Generative Models}
\begin{figure*}
\centering     
\subfigure[Encoding and Tokenization]{\label{fig:bd-a}\includegraphics[width=0.5\textwidth]{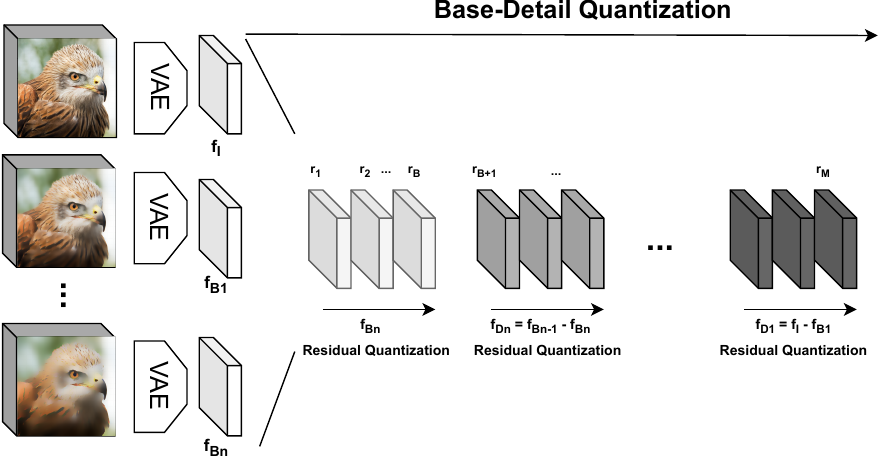}}
\subfigure[Next Detail Prediction]{\label{fig:bd-b}\includegraphics[width=0.37\textwidth]{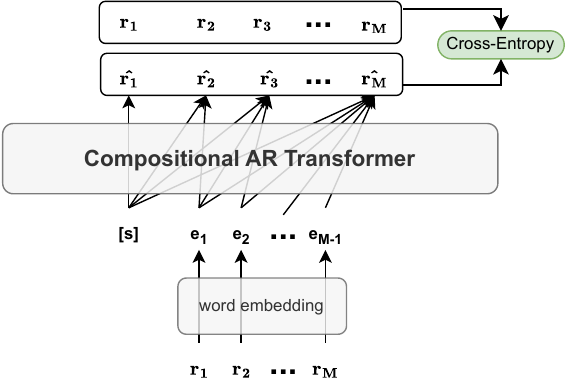}}

\caption{Overview of the CART Approach.}
\end{figure*}
Generative models for image synthesis have advanced rapidly, enabling both unconditional and conditional generation based on priors. VAEs \cite{kingma2013auto, shao2020controlvae} and GANs \cite{goodfellow2020generative, mirza2014conditional} established foundational approaches, with GANs generating high-quality images via adversarial training.
Diffusion models \cite{ho2020denoising, song2020denoising, ho2022cascaded} introduce sequential denoising processes, gradually refining noise into realistic images. Their remarkable ability to synthesize high quality images with fine-grained visual details have enabled applications in text-to-image generation \cite{zhang2023text,zhu2023conditional}, inpainting \cite{lugmayr2022repaint, corneanu2024latentpaint,yang2023uni}, super-resolution \cite{yue2024resshift,li2022srdiff}, 3D reconstruction \cite{anciukevivcius2023renderdiffusion, zhou2023sparsefusion}, and image editing \cite{brooks2023instructpix2pix, kawar2023imagic, bala2024galaxyedit}. However, their many iterative steps add computational overhead, limiting scalability for real-time, high-resolution synthesis.

\subsection{Auto-Regressive Generative Models}

Auto-Regressive (AR) models attempt to predict next tokens in a sequence while conditioned on previous tokens. GPT models \cite{brown2020language, radford2019language} using transformers \cite{vaswani2017attention} achieved revolutionary success in language tasks, motivating computer vision applications. Early attempts included DRAW \cite{gregor2015draw} with sequential variational auto-encoding using RNNs, and pixel-level prediction approaches (PixelCNN \cite{salimans2017pixelcnn++}, PixelRNN \cite{van2016pixel}, and Image Transformer \cite{parmar2018image}). However, sequentially predicting billions of pixels proved computationally prohibitive, and Image-GPT \cite{chen2020generative} with 6.8B parameters only achieved image generation at $96\times96$ resolution.
Vector Quantized VAE (VQ-VAE) \cite{van2017neural} addressed scalability by compressing images into discrete token sequences. In \cite{parmar2018image} transformer decoder was utilized to enable AR generation using VQ-VAE tokens. VAR \cite{tian2024visual} demonstrated that token ordering critically impacts AR image generation, and proposed multi-scale tokenization with “next-scale" prediction.

\subsection{Vector Quantized VAE (VQ-VAE)}

In order to perform AR modeling of images via next-token prediction, VQ-VAE is utilized to tokenize the image into discrete tokens. The encoder $\bm{\mathcal{E}}$, converts images to feature maps $\bm{f} = \bm{\mathcal{E}}(\bm{I}) \in \mathbb{R}^{h \times w \times C}$, followed by quantization to discrete tokens $\bm{q} = \bm{\mathcal{Q}}(\bm{f}) \in [V]^{h \times w}$ using learnable codebook $\mathcal{Z} \in \mathbb{R}^{V \times C}$ with $V$ vectors,

\begin{equation}
    q^{(i,j)} = \big( arg \min_{v \in [V]} || \text{look-up}(\mathcal{Z}, v) - f^{(i,j)} ||_2 \big) \in [V],
\end{equation}
where $\text{look-up}(\mathcal{Z},v)$ refers to taking the $v^{th}$ vector in codebook $\mathcal{Z}$. Reconstruction involves codebook lookup $\hat{\bm{f}} = \text{look-up}(\mathcal{Z}, \bm{q})$ and decoding $\hat{\bm{I}} = \bm{\mathcal{D}}(\hat{\bm{f}})$.
Training a VQ-VAE involves the minimization of a compound loss,
\begin{equation}
\label{eq:compound_loss}
    || \bm{I}-\hat{\bm{I}} ||_2 + || \bm{f}-\hat{\bm{f}} ||_2 + \lambda_p \mathcal{L}_p(\hat{\bm{I}}) + \lambda_G \mathcal{L}_G(\hat{\bm{I}}),
\end{equation}
where $\mathcal{L}_p$ is perceptual loss (LPIPS \cite{zhang2018unreasonable}), $\mathcal{L}_G$ is discriminative loss (StyleGAN \cite{karras2019style}), and $\lambda_p$ and $\lambda_G$ are the corresponding loss weights.

\subsection{Mumford-Shah Functional}



The Mumford-Shah functional \cite{mumford1989optimal} provides a form of all regularizers aiming at discontinuity-preserving smoothing given a bounded set $\Omega \in \mathbb{R}^d$,


\begin{equation}
    \min_{u,K}  \int_{\Omega} |u-f|^2 dx + \alpha \int_{\Omega / K} |\nabla u|^2 dx + \lambda |K|,
\end{equation}
This approximates vector-valued input image $f : \Omega \to \mathbb{R}^k$ with function $u: \Omega \to \mathbb{R}^k$, which is smooth everywhere except at $(d-1)$-dimensional jump set $K$. $\lambda>0$ controls the length of $K$. A common approach to solve the Mumford-Shah functional is the Ambrosio-Tortorelli approach \shortcite{ambrosio1990approximation},


\begin{equation}
\label{eq:AT}
\begin{split}
    \min_{u,s} \int_{\Omega} |u-f|^2 dx + \alpha \int_{\Omega} (1-s)^2 |\nabla u|^2 dx \\ 
    + \lambda \int_{\Omega} (\epsilon |\nabla s|^2 + \frac{1}{4\epsilon} s^2) dx,
\end{split}
\end{equation} 

with a small parameter $\epsilon > 0$ and an edge set indicator $s: \Omega \to \mathbb{R}$. The points $x \in \Omega$ are part of the edge set $K$ if $s(x) \approx 1$ and part of smooth region if $s(x) \approx 0$. The variables $u$ and $s$ are found by alternating minimization.
 


 \section{Proposed Approach}
\begin{figure*}
\centering
    \includegraphics[width=0.75\textwidth]{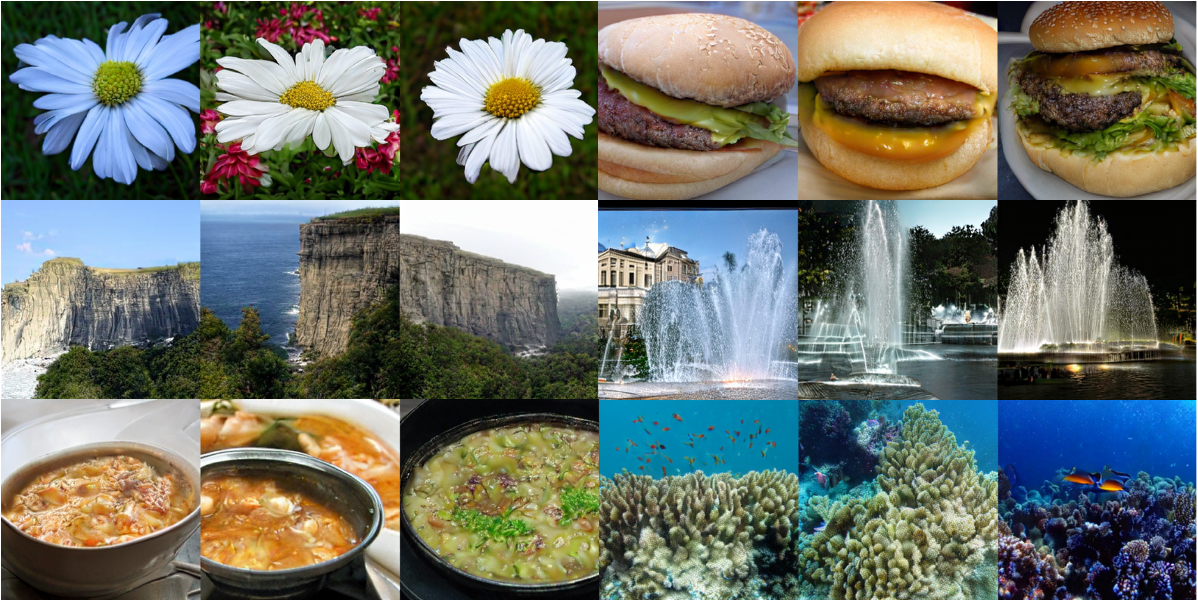}
    \caption{Generated Samples using CART-256}
    \label{fig:CART_Results}

\end{figure*}
We propose a novel approach for autoregressive image generation where the model initially generates a base image focusing on global structure, and subsequently refines it through iterative detail addition. Our training methodology comprises three steps: \textbf{(1) Decomposition}: Each training image is decomposed into $n$ hierarchical factors representing progressive detail layers, \textbf{(2) Encoding and Tokenization}: The factors are encoded into a latent space using a VQ-VAE, preserving essential features while reducing dimensionality, \textbf{(3) Iterative Prediction}: A Transformer decoder is trained to predict successive detail factors (token-maps), enabling incremental detail addition.
\begin{figure}[ht]
    \centering
    \includegraphics[width=0.35\textwidth]{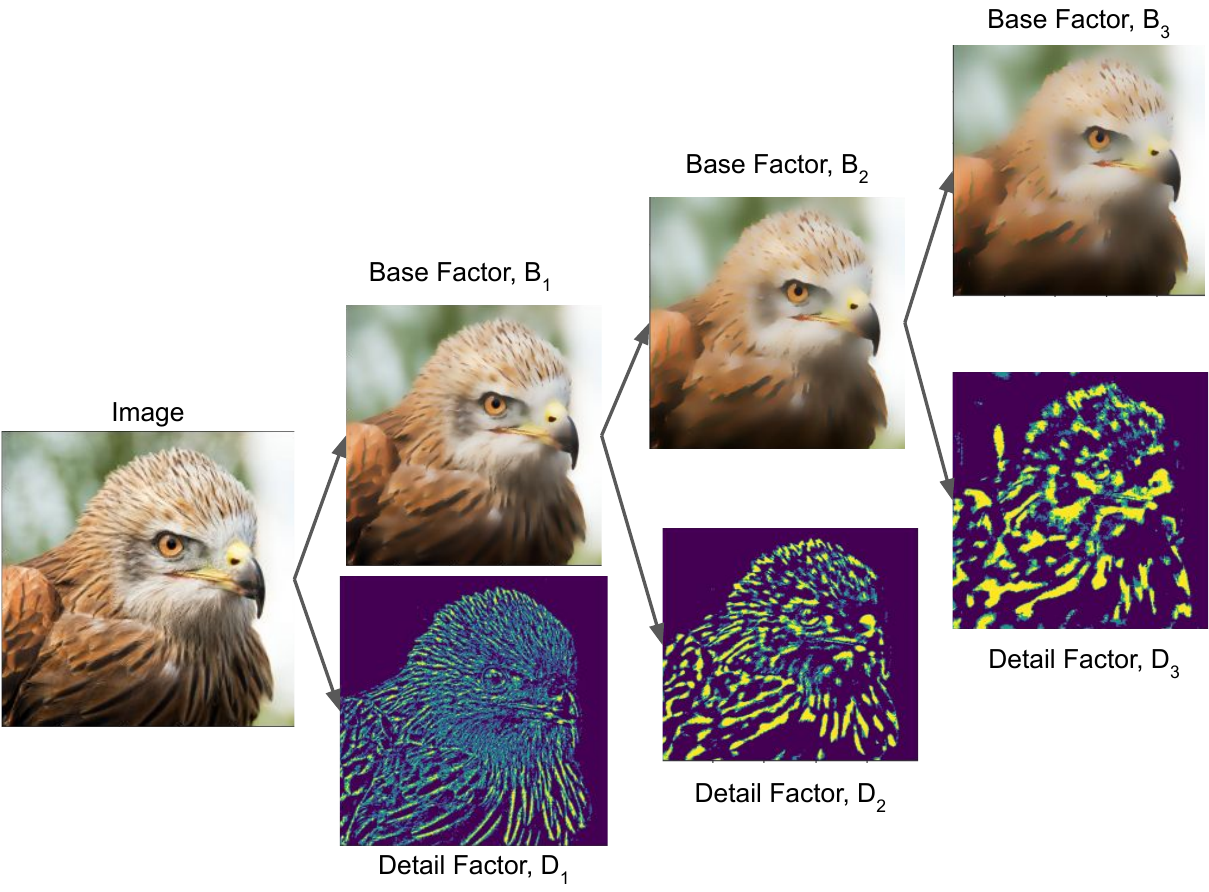}
    \caption{Hierarchical Base-Detail Decomposition}
    \label{fig:decomp}

\end{figure}

\subsection{Hierarchical Base-Detail Decomposition}
\label{sec:hirarchical_bd}


An image can be represented as a linear combination of factor images capturing distinct properties of the image. We decompose an image into a base and a detail factor,
\begin{equation}
\bm{I} = \bm{B} + \bm{D}, 
\label{eq:base-detail-decomp}
\end{equation}
where $\bm{I}, \bm{B}, \bm{D} \in \mathbb{R}^{H \times W \times 3}$ denote a training image and its corresponding base and detail factors. The base factor $\bm{B}$ is obtained by minimizing the Mumford-Shah functional via the Ambrosio-Tortorelli approach, as detailed in Eq. \ref{eq:AT}.
This base factor can be recursively decomposed to yield multiple detail factors, 
\begin{equation}
\label{eq:nth_order_decomp}
    \bm{I} = \bm{B}_n + \bm{D}_n + \bm{D}_{n-1} + ... + \bm{D}_1,
\end{equation}
where, $\bm{B}_{k-1} = \bm{B}_k + \bm{D}_k, \text{ } \forall k \in \{ 1,...,n \}$. Equation \ref{eq:nth_order_decomp} defines the $n^{th}$ order decomposition of $\bm{I}$. In this decomposition, the base factor $\bm{B}_n$ captures the image’s overall structure, composition, and global features, while the detail factors $\{\bm{D}_k\}_{k=1}^n$ represent local features that contribute to the finer details of the image. Figure \ref{fig:decomp} shows the hierarchical base-detail decomposition process. 

We adopt edge-aware smoothing over frequency-based decomposition methods to preserve structural integrity in base images. While frequency-domain approaches such as Discrete Cosine Transform (DCT) \cite{nash2021generating} and Wavelet Transforms \cite{yu2021wavefill} provide computational efficiency, they exhibit fundamental limitations for our compositional framework. DCT-based decomposition applies uniform smoothing across both global structures and local features, failing to distinguish between semantically important edges and fine-grained textures. Additionally, the inverse DCT/Wavelet transformation introduces ringing artifacts that compromise image quality in the reconstructed base component.
In contrast, Mumford-Shah smoothing provides selective regularization that preserves global edges and large-scale structural elements while effectively smoothing textural and local features. This edge-preserving property enables successful disentanglement of structural information (captured in base factors) from fine-grained details (captured in detail factors), which is critical for our iterative refinement approach (Further discussion in supplement. 
Our framework maintains flexibility by supporting various image decomposition techniques within the general formulation of Equation \ref{eq:nth_order_decomp}. 
This modularity allows for domain-specific decomposition strategies while preserving the core auto-regressive generation mechanism. 


\subsection{Encoding and Tokenization}
\begin{figure*}
\centering
    \includegraphics[width=0.85\textwidth]{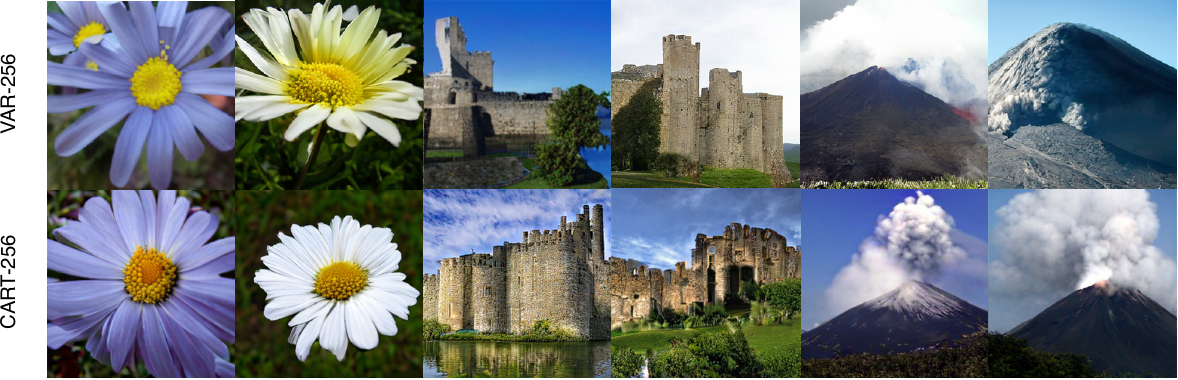}
    \caption{Comparison of samples generated by VAR-256 (top) and CART-256 (bottom).}
    \label{fig:CARTvsVAR}
\end{figure*}



In our approach, each image is represented by token maps $\{r_1, r_2, ...,r_\mathcal{M}\}$ within the latent space of a Vector Quantized Variational AutoEncoder (VQ-VAE), rather than single tokens. This token-map representation preserves the spatial coherence of the feature map and reinforces the spatial structure inherent in the image. Departing from the multi-scale approach in VAR \cite{tian2024visual}, we propose a tokenization scheme such that these token maps represent the base and detail factors. 
Specifically, the image representation is comprised of $\mathcal{B}$ base token maps, $(r_1,...,r_{\mathcal{B}}), \text{ where } \mathcal{B} < \mathcal{M}$ and ($\mathcal{M}-\mathcal{B}$) detail token maps, $(r_{\mathcal{B}+1}, ..., r_{\mathcal{M}})$.

Following the Base-Detail Decomposition, we encode the original image $I$ along with the Base Factors $\{\bm{B}_k\}_{k=1}^{n}$ using a VAE, 

\begin{equation}
    \bm{f}_{\bm{B}_k} = \bm{\mathcal{E}}(\bm{B}_k),
\end{equation}

where $\bm{f}_{\bm{B}_k} \in \mathbb{R}^{h \times w \times c} \text{  } \forall k \in \{1,...,n\}$. 
The token maps representing the base factor $\bm{B}_n$, $\{ r_1,...,r_{\mathcal{B}} \}$ are created by performing residual quantization \cite{lee2022autoregressive} on the encoded feature map $\bm{f}_{\bm{B}_n}$ with a quantization depth of $\mathcal{B}$.
The encoded representation of the $k^{th}$ detail factor is then determined as follows,

\begin{equation}
    \bm{f}_{\bm{D}_k} = \bm{f}_{\bm{B}_{k-1}} - \bm{f}_{\bm{B}_k},
\end{equation}

where, $\bm{f}_{\bm{B}_k}$ is the encoded representation of the $k^{th}$ base factor and $\bm{f}_{\bm{B}_0} = \bm{f}_{\bm{I}}$. Each detail factor is quantized with quantization depth $\frac{(\mathcal{M}-\mathcal{B})}{n}$ , yielding the remaining tokens, as illustrated in Figure \ref{fig:bd-a}. The full algorithm for extracting token maps from a given image is presented in Algorithm \ref{alg:encode_main}. 

\subsection{Iterative Detail Learning}

We employ an auto-regressive approach to predict each successive “next-detail" token map. 
Given the set of tokens $\{r_1,r_2,...,r_{\mathcal{M}}\}$, the autoregressive likelihood is defined as, 
\begin{equation}
    P(r_1, ..., r_{\mathcal{M}}) = \prod_{m=1}^{\mathcal{M}} P(r_m|r_1,...,r_{m-1}).
\end{equation}
where each autoregressive unit, $r_m \in [V]^{h_m \times w_m}$ is a token map containing $h_m \times w_m$ tokens. 

For the model architecture, we utilize a standard decoder-only Transformer architecture similar to that in GPT-2 \cite{radford2019language}, VQ-GAN \cite{esser2021taming}, and VAR \cite{tian2024visual}. At each auto-regressive step, the Transformer decoder predicts the distribution over all $h_m \times w_m$ tokens in parallel as depicted in Figure \ref{fig:bd-b}. To enforce causality, we apply a causal attention mask, ensuring that each token map $r_m$ only attends to its preceding tokens $r_{\leq m}$.




\section{Experiments}

\subsection{Implementation Details}

\begin{algorithm}    

    \caption{Base-Detail VQ-VAE Encoding}
    \label{alg:encode_main}

    \textbf{Input:} 
    \begin{itemize}
        \item Raw image, $I$
        \item Target Image dimensions, $h_{\mathcal{M}}, w_{\mathcal{M}}$
        \item Base Image, $B_n$
    \end{itemize}

    \textbf{ }
    
    \textbf{Hyperparameters: } 
    \begin{itemize}
        \item Total number of tokens to represent the image, $\mathcal{M}$
        \item number of base tokens, $\mathcal{B}$
        \item number of detail factors, $n$
    \end{itemize}

    \textbf{ }
    
    \SetKwBlock{Beginn}{beginn}{ende}

    \Begin{
        $f_I \gets \varepsilon(I)$\
        
        $f_B \gets \varepsilon(B)$\ 
        
        $f_{D_i} \gets f_{B_i-1} - f_{B_i} \text{ } \forall i \in \{1,...,n\}$\ 
        
        $t \gets -1$\ 
        
        \For {k=1:$\mathcal{M}$}
            {
            \If{$k \leq \mathcal{B}$}
                {
                \text{ }
                
                $r_k \gets \mathcal{Q}(\text{interpolate}(f_B, h_k, w_k))$\ 
                
                $R \gets queue_{push}(R,r_k)$\ 
                
                $z_k \gets \text{LookUp}(r_k, \mathcal{Z})$\ 
                
                $z_k \gets \text{interpolate}(z_k, h_{\mathcal{M}}, w_{\mathcal{M}})$\ 
                
                
                $f_B \gets f_B - \phi_k(z_k)$\

                \text{ }
                }
            \Else 
              {  
              \If{mod($k$, $\frac{\mathcal{M}-\mathcal{B}}{n}$) $= 0$}
                   { $t \gets t+1$}
                
                $r_k \gets \mathcal{Q}(f_{D_t})$\ 
                
                $R \gets queue_{push}(R,r_k)$\ 
                
                $z_k \gets \text{LookUp}(r_k, \mathcal{Z})$\ 
                
                
                $f_{D_t} \gets f_{D_t} - \phi_k(z_k)$\ 
                
            }
        }
        \Return base-detail tokens $R$.\ 
        
    }

\end{algorithm}

For detail decomposition of training images, Mumford-Shah smoothing (Equation \ref{eq:AT}) with $\alpha=1$, $\lambda=0.01$ is used. Each training image is decomposed iteratively to obtain a $3^{rd}$ order decomposition, $\bm{I} = \bm{B}_3 + \bm{D}_3 + \bm{D}_2 + \bm{D}_1$. 
Note that the computational overhead due to Mumford-Shah decomposition is a one-time cost, as the decomposition is only utilized during the training process and not during inference. 
A Vanilla VQ-VAE \cite{van2017neural} is used along with $\mathcal{M}$ extra convolutions to realize the Base-Detail quantization scheme as depicted in Figure \ref{fig:bd-a} and Algorithm \ref{alg:encode_main}. To mitigate information loss when upscaling $z_k$ to the target resolution $h_{\mathcal{M}} \times w_{\mathcal{M}}$, we introduce an additional set of $\mathcal{M}$ convolutional layers, denoted as $\{ \phi_k \}_{k=1}^{\mathcal{M}}$ which enhance feature refinement and preserve structural details. The base and detail factors share the same code book with $V=4096$. As in \cite{tian2024visual,esser2021taming}, the tokenizer is trained on OpenImages \cite{kuznetsova2020open} with Compound loss (Equation \ref{eq:compound_loss}) and spatial downsample of $16 \times$. 

\begin{figure*}
\centering
    \includegraphics[width=0.99\textwidth]{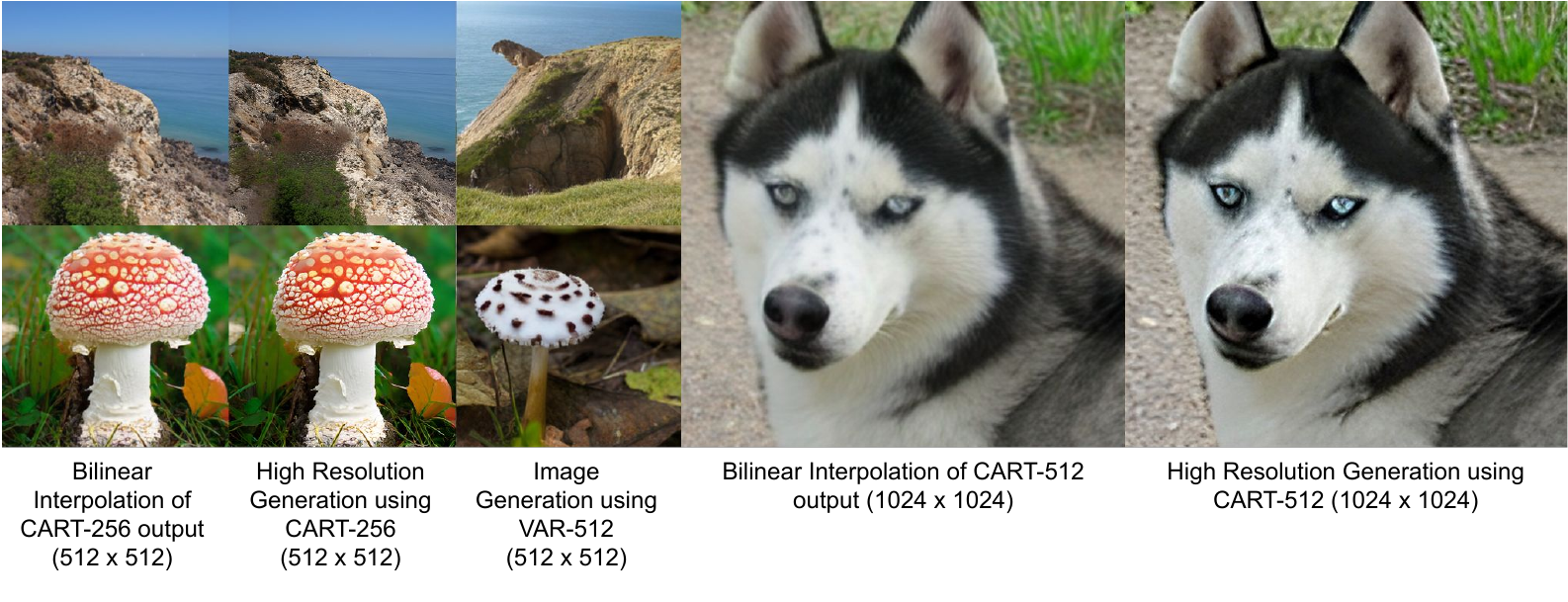}
    \caption{High Resolution image generation using patchwise detail prediction using CART-d30-256 and CART-d30-512. Zoom-in recommended to observe finer details. }
    \label{fig:bilinearvsCART}

\end{figure*}

\begin{algorithm}
\caption{Base-Detail VQ-VAE Reconstruction} \label{alg:decode_main}

    \textbf{Input:} Base-Detail Tokens, $R$.
    
    \textbf{Hyperparameters: } 
    \begin{itemize}
        \item Total number of tokens to represent the image, $\mathcal{M}$
        \item number of base tokens, $\mathcal{B}$
        \item number of detail factors, $n$
    \end{itemize}

    \SetKwBlock{Beginn}{beginn}{ende}
    
    \Begin{
        $\hat{f} \gets 0$\  
        
        \For {k=1:$\mathcal{M}$}{
            \If{ $k \leq \mathcal{B}$}{
                $r_k \gets queue_{\text{pop}}(R) $\
                $z_k \gets lookup(Z,r_k)$\
                $z_k \gets \text{interpolate}(z_k, h_k, w_k)$\
                $\hat{f} \gets \hat{f} + \phi_k(z_k)$\
                }
            \Else{
                $r_k \gets queue_{\text{pop}}(R) $\
                $z_k \gets lookup(Z,r_k)$\
                $\hat{f} \gets \hat{f} + \phi_k(z_k)$\
                }}

        $\hat{I} \gets \mathcal{D}(\hat{f})$\
        
        \Return reconstructed image $\hat{I}$\
    }
\end{algorithm}

The tokenized base-detail factors are then utilized to train a Transformer Decoder architecture which learns to predict the “next-detail" token. A standard decoder-only transformer architecture is used similar to GPT-2 \cite{radford2019language} and VQGAN \cite{esser2021taming}. We use a total of 14 steps to generate an image, including 8 steps to generate the base factor and 6 steps to generate the detail factors. During inference, the Transformer predicts the codes and the VQ-VAE decoder decodes the generated image. The decoding rocess is summarized in Algorithm \ref{alg:decode_main}. The depth of the transformer is varied from 16 to 30 to obtain models with varying complexity. The model is trained with initial learning rate of $1e^{-4}$. For training, we use 16 A100 GPUs with a global batch size of 768 for CART-d30-256 and batch size of 384 for CART-d30-512. All visual results are generated with seed 42 and quantitative results are averaged over 10 randomly selected seeds.

\subsection{Emperical Results}





\begin{table*}
\centering
    \begin{tabular}{c|c|c|c|c|c}
        \hline
        Type & Model & FID $\downarrow$ & IS $\uparrow$ & Params & Steps  \\
        \hline
        GAN & BigGAN \cite{brock2018large} &6.95 & 224.5 & 112M & 1 \\
        GAN & GigaGAN \cite{kang2023scaling} & 3.45 & 225.5 & 569M & 1 \\
        GAN & StyleGAN-XL \cite{sauer2022stylegan} & 2.30 & 265.1 & 166M & 1 \\
        \hline
        Diffusion & ADM \cite{dhariwal2021diffusion} & 10.94 & 101.0 & 554M & 250\\
        Diffusion & CDM \cite{ho2022cascaded} & 4.88 & 158.7 & - & 8100\\
        Diffusion & LDM-4-G \cite{rombach2022high} & 3.60 & 247.7 & 400M & 250\\
        Diffusion & DiT-XL/2 \cite{peebles2023scalable} & 2.27 & 278.2 & 675M & 250\\
        Diffusion & L-DiT-3B \cite{LargeDiT} & 2.10 & 304.4 & 3.0B & 250\\
        Diffusion & L-DiT-7B \cite{LargeDiT} & 2.28 & 316.2 & 7.0B & 250\\
        Diffusion & DiffiT \cite{hatamizadeh2024diffit} & 1.73 & 276.5 & 561M & 250 \\
        \hline
        Mask & MaskGIT \cite{chang2022maskgit} & 6.18 & 182.1 & 227M & 8\\
        Mask & RCG \cite{li2023self} & 3.49 & 215.5 & 502M & 20\\
        \hline
        AR & VQVAE-2 \cite{razavi2019generating} & 31.11 & - & 13.5B & 5120 \\
        AR & DCTransformer \cite{nash2021generating} & 36.51 & - & 738M & - \\
        AR & VQGAN-re \cite{esser2021taming} & 5.20 & 280.3 & 1.4B & 256 \\
        AR & ViTVQ-re \cite{yu2021vector} & 3.04 & 227.4 & 1.7B & 1024 \\
        AR & RQTran-re \cite{lee2022autoregressive} & 3.80 & 323.7 & 3.8B & 68 \\
        AR & LlamaGen \cite{sun2024autoregressive} & 2.18 & 263.3 & 3.1B & 576 \\
        AR & SpectralAR-d24 \cite{huang2025spectralar} & 2.13 & 307.7 & 1.0B & 64 \\
        \hline
        VAR & VAR-d16 \cite{tian2024visual} & 3.30 & 274.4 & 310M & 10 \\
        VAR & VAR-d30 \cite{tian2024visual} & 1.92 & 323.1 & 2B & 10 \\
        VAR & VAR-d30-re \cite{tian2024visual} & 1.73 & 350.2 & 2B & 10 \\
        VAR & VAR-d30-re \cite{tian2024visual} & 1.70 & 352.8 & 2B & 14 \\
        \hline
        CART & CART-d16 & 2.89 & 293.0 & 310M & 14\\
        CART & CART-d24 & 1.90 & 328.1 & 1.0B & 14\\
        CART & CART-d24-re & 1.77 & 345.7 & 1.0B & 14 \\
        CART & CART-d30 & 1.65  & 366.8 & 2.0B & 14 \\
        CART & CART-d30-re & 1.61  & 377.5 & 2.0B & 10 \\
        CART & CART-d30-re & \textbf{1.57}  & \textbf{381.9} & 2.0B & 14 \\
        \hline
         & (val. data) & 1.78 & 236.9 & & \\
        \hline

    \end{tabular}
    \caption{Quantitative results on ImageNet $256 \times 256$. Suffix '-re' refers to models that use rejection sampling}
    \label{tab:IN256}
\end{table*}

The proposed CART model was evaluated on ImageNet \cite{deng2009imagenet} at $256 \times 256$ (CART-256) and $512 \times 512$ (CART-512) resolutions for benchmarking against SOTA image generation methods. Comparative results in Tables \ref{tab:IN256} and \ref{tab:IN512} show that CART outperforms SOTA AR and Diffusion models and achieves FID lower than ImageNet validation set while maintaining comparable complexity and generation steps. CART benefits from base-detail decomposition that disentangles global structures from local details, simplifying the learning process through more natural token ordering. Figure \ref{fig:CART_Results} depicts generated images using our method, while Figure \ref{fig:CARTvsVAR} compares VAR \cite{tian2024visual} and CART outputs. CART produces images with enhanced details and structure compared to VAR's “next-scale" prediction scheme. CART surpasses both Diffusion Transformer \cite{peebles2023scalable, LargeDiT, hatamizadeh2024diffit} and SOTA VAR \cite{tian2024visual} in autoregressive image generation. 
Extended results are provided in the supplement. 

\begin{figure*}
\centering
    \includegraphics[width=0.8\textwidth]{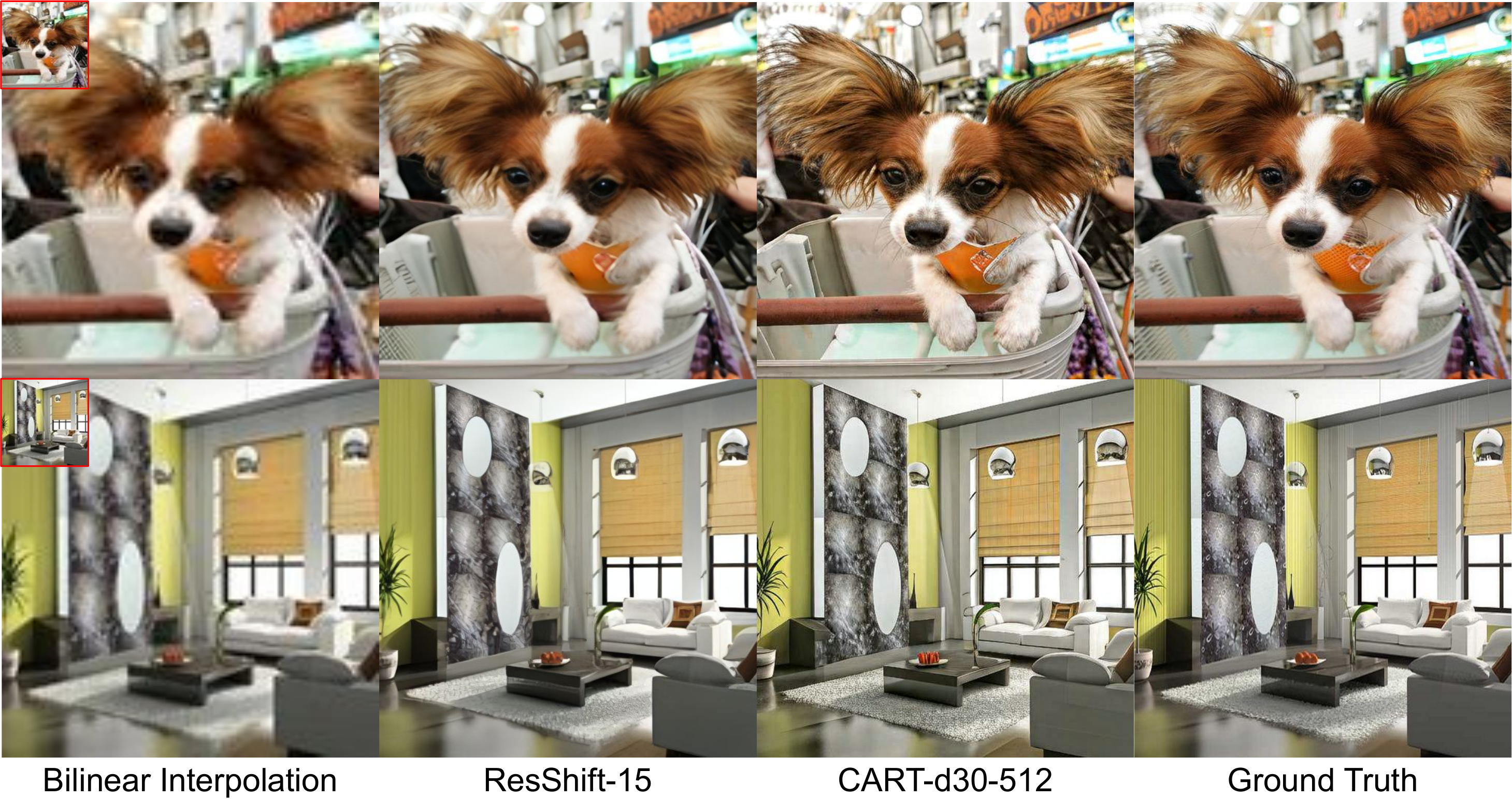}
    \caption{Comparison of CART for super-resolution with ResShift. Zoom-in recommended to observe finer details.} 
    \label{fig:sr1}

\end{figure*}

\begin{figure*}
\centering     
\subfigure[Recoloring of generated images by manipulating albedo]{\label{fig:recolor1}\includegraphics[width=0.6\textwidth]{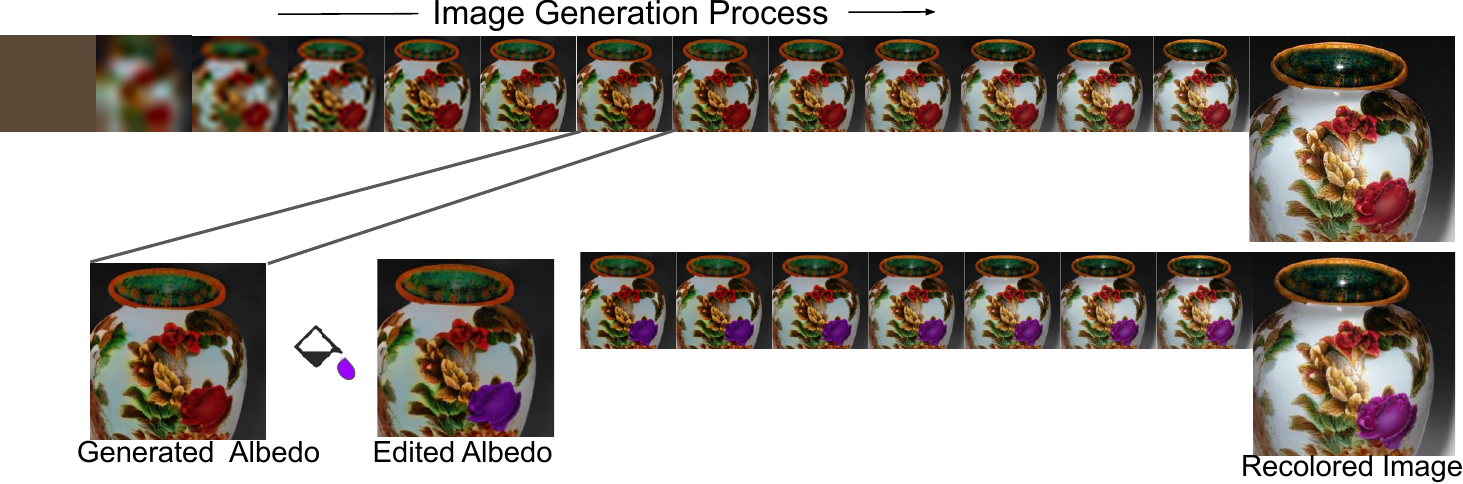}}
\subfigure[Recoloring of “bell-pepper" and “tent" classes]{\label{fig:recolor2}\includegraphics[width=0.39\textwidth]{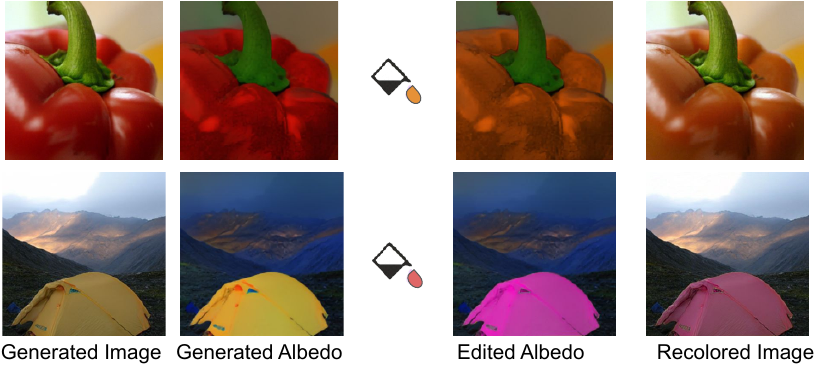}}

\caption{Recoloring of generated images when using intrinsic decomposition to tokenize the image.}
\end{figure*}

 \begin{table}[]
        \centering
        \begin{tabular}{c|c|c|c}
        \hline
            Type & Model & FID $\downarrow$ & IS $\uparrow$\\
            \hline
             GAN & BigGAN \shortcite{brock2018large} & 8.43 & 177.9 \\
             \hline
             Diff. & ADM \shortcite{dhariwal2021diffusion} & 23.24 & 101.0 \\
             Diff. & DiT-XL/2 \shortcite{peebles2023scalable} & 3.04 & 240.8\\
             \hline
             Mask & MaskGiT \shortcite{chang2022maskgit} & 7.32 & 156.0 \\
             \hline
             AR & VQGAN \shortcite{esser2021taming} & 26.52 & 66.8 \\
             \hline
             VAR & VAR-d36-s \shortcite{tian2024visual} & 2.63 & 303.2 \\
             \hline
             CART & CART-NOV-256-d30 & 2.85 & 297.1 \\
             CART & CART-OV-256-d30 & \underline{2.54} & \underline{305.7} \\
             CART & CART-512-d30 & \textbf{2.40} & \textbf{315.5} \\
             \hline
             
        \end{tabular}
        \caption{Quantitative results on ImageNet $512 \times 512$.}
        \label{tab:IN512}
    \end{table}



\subsection{Other Applications}
\subsubsection{Generalizing to Higher Resolutions}
\label{sec:high_res}

A key advantage of employing base-detail decomposition is the explicit disentanglement of global and local image features, facilitating high-resolution image synthesis and image super-resolution even when trained on lower-resolution inputs. Empirically, we observe that the base factor encapsulates global attributes, including class-conditional structure and overall color composition, while the detail factor captures local features such as textures and fine-grained details (see Figure \ref{fig:cartvvar_main}). This decomposition allows the base factor to be upscaled without loss of essential global information, while the detail factor is generated in a patchwise manner. Since the detail factor inherently lacks dependencies on global structures, patchwise synthesis does not introduce any discontinuities. Figure \ref{fig:bilinearvsCART} compares bilinear upscaling and VAR \cite{tian2024visual} with our method, demonstrating that reusing lower-resolution base images and introducing patchwise details at target resolution effectively preserves content while enhancing fine details. Table \ref{tab:IN512} presents performance comparisons against state-of-the-art methods. “CART-256-NOV" refers to non-overlapping patchwise detail generation at 512×512, while “CART-256-OV" employs 50\% overlapping patches for improved continuity. “CART-512" corresponds to full training on 512×512 images. Notably, CART (trained at 256×256) outperforms VAR models trained from scratch at 512×512. Further details and results are provided in the supplement. 



\subsubsection{Super-Resolution (SR)}

Given a Low-Resolution (LR) image, we encode it using Base-Detail VQVAE token maps $\{r_1,...,r_k\}$, where $r_k$ has resolution $h/p \times w/p$ ($h$, $w$ are LR image dimensions, $p$ is the down-sampling factor of VQ-VAE). $\{r_1,...,r_k\}$ are appended as past tokens to CART's prediction sequence, which then predicts subsequent token maps $\{r_{k+1},...,r_{\mathcal{M}}\}$ in unconditional generation setting. For target resolutions exceeding training resolution, we apply the high-resolution generation strategy described above.
Table \ref{tab:INSR} compares our SR results with SOTA generative methods specifically trained or fine-tuned for super-resolution tasks and Figure \ref{fig:sr1} provides visual comparison. 
Although CART yields lower PSNR than ResShift \shortcite{yue2024resshift}, it surpasses all competing methods in CLIP-IQA score \cite{wang2023exploring}, indicating superior perceptual image quality as assessed by human visual preference. Extended results are provided in the supplement. 

\begin{figure}
\centering
    \includegraphics[width=0.49\textwidth]{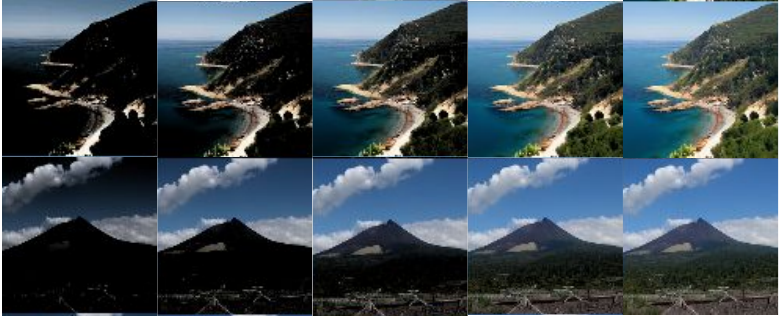}
    \caption{Generation using specularity decomposition. First Column: Base Diffuse Term, $2^{nd}$-$5^{th}$ Column: addition of specular terms to change global illumination.}
    \label{fig:spec3}

\end{figure}

\begin{table}[]
        \centering
        \begin{tabular}{c|c|c|c}
        \hline
            Model & PSNR $\uparrow$ & SSIM $\uparrow$ & CLIPIQA $\uparrow$ \\
            \hline
             Real-ESRGAN \shortcite{wang2021real}& 24.04 & 0.665 & 0.523 \\
             \hline
             ResShift-15 \shortcite{yue2024resshift} & \textbf{24.90}  & \textbf{0.673} & 0.603\\
             Sin-SR \shortcite{wang2024sinsr} & 24.56 & 0.657 &  0.611 \\
             \hline
             CART-256-d30 & 24.16 & 0.633 & 0.594\\
             CART-512-d30 & \underline{24.65} & \underline{0.660} & \textbf{0.672}\\
             \hline
             
        \end{tabular}
        \caption{Comparison of CART models with specialized Super-Resolution models. Metrics are reported for SR from $128 \times 128$ to $512 \times 512$ resolution on ImageNet Test Set.} 
        \label{tab:INSR}
    \end{table}

\subsubsection{Recoloring Generated Images}


Intrinsic image decomposition \cite{careaga2023intrinsic} provides a principled approach to disentangle images into reflectance (albedo) and illumination (shading) components, enabling semantically meaningful manipulations. CART adopts this decomposition during training to encourage controllable generation. The observed image $\bm{I}$ is modeled as the composition of albedo and shading map, $\bm{I} = \bm{A}\star\bm{S}$. Where $\bm{A} \in \mathbb{R}^{H \times W \times 3}$ encodes illumination-invariant properties (object color and structure) and $\bm{S} \in \mathbb{R}^{H \times W \times 1}$ captures illumination-dependent effects.
To facilitate learning and component-wise manipulation, we convert the multiplicative decomposition to additive form via logarithmic transformation, $\log \bm{I} = \log \bm{A} + \log \bm{S}$.
CART leverages this formulation by learning to predict the log-image and reconstructing via exponentiation, $\bm{I} = \exp(\log \bm{A} + \log \bm{S})$. Images are tokenized into 14 steps comprising 7 albedo and 7 shading token maps. This layered approach enables explicit learning of color and lighting factors. The decomposition and separate supervision apply only during training to induce generative factor separation. At inference, CART directly generates compositional outputs without explicit decomposition. By structurally separating these factors during training, CART supports controllable color and illumination in generated images while compositional constraints ensure globally coherent synthesis. Figure \ref{fig:recolor1} depicts the process of image generation and color manipulation using this decomposition. 
Figure \ref{fig:recolor2} depicts more instances of recoloring.


\subsubsection{Lighting Control of Generated Images}

Replacing base-detail decomposition in Equation \ref{eq:base-detail-decomp} with Specularity decomposition \cite{saini2024specularity} enables explicit lighting control in generated images. Following the dichromatic reflection model \cite{tominaga1994dichromatic}, images consist of diffuse $(\bm{A})$ and specular $(\bm{E})$ components: $\bm{I} = \bm{A} + \bm{E}$. We employ $4^{th}$-order decomposition for four illumination control levels: $\bm{I} = \bm{A}_4 + \bm{E}_4 + \bm{E}_3 + \bm{E}_2 + \bm{E}_1$. Generation uses 16 autoregressive steps: 8 for base factor generation and 8 for controlled lighting refinement. Figure \ref{fig:spec3} demonstrates synthesized images with varying illumination while maintaining structural consistency for classes “cliff" and “volcano".  Extended results are provided in the supplement. 


\subsection{Ablation Study }
\begin{figure}
\centering
    \includegraphics[width=0.49\textwidth]{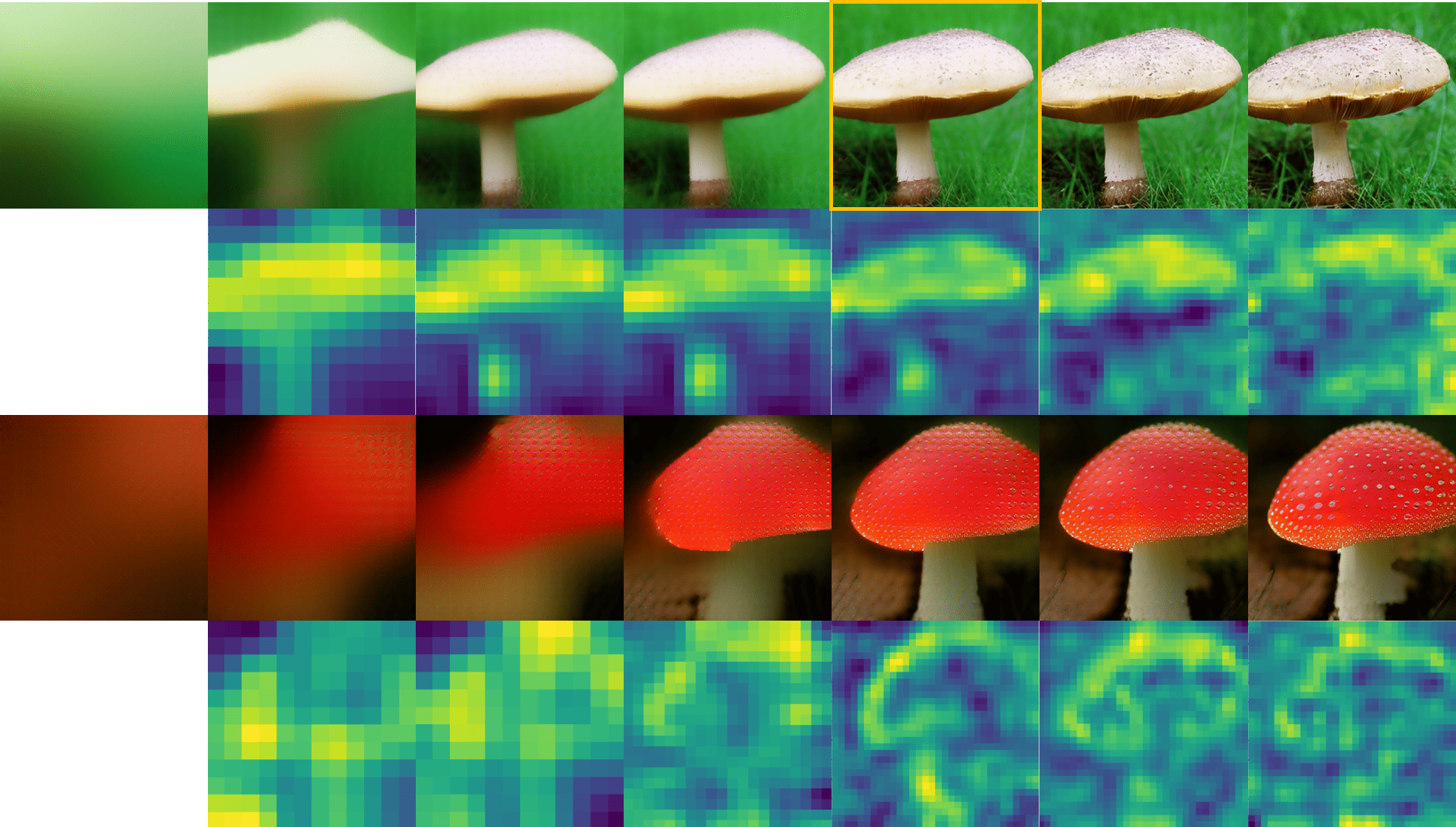}
    \caption{Top row: Intermediate visual results for CART. Base image is marked by yellow outline. $2^{nd}$ row: Self-attention maps for corresponding intermediate layers of CART. $3^{rd}$ row: Intermediate visual results for VAR. bottom row: Self-attention maps for corresponding intermediate layers of VAR.}
    \label{fig:cartvvar_main}

\end{figure}
Table \ref{tab:ablation1} evaluates the impact of various CART model components. While employing multi-scale tokenization for the base factor yields only marginal gains in FID, this approach significantly reduces memory usage and accelerates generation, offering practical advantages for larger models. Table \ref{tab:dec_order} compares CART performance across different decomposition orders. Decomposition order $0$ is equivalent to VAR. Best performance occurs with $3^{rd}$ order Base-Detail decomposition. Beyond $3^{rd}$ order, the base image becomes over-smoothed and loses essential global structural details, leading to sub-optimal learning. 

\begin{table}[]
    \centering
    \begin{tabular}{c|c|c|c|c}
    \hline
        Model & CFG & \makecell{MS \\ Tokens} & \makecell{BD \\ Tokens} & FID \\
        \hline
         AR \shortcite{esser2021taming} & \xmark & \xmark & \xmark & 18.65 \\ 
         \hline
         \makecell{VAR-d16 \shortcite{tian2024visual}} & \cmark & \cmark & \xmark & 3.30\\
         \hline
         CART-d16 & \cmark & \xmark & \cmark & 2.90\\
         CART-d16 & \cmark & \cmark & \cmark & 2.89\\ 
         \hline
         \makecell{VAR-d30 \shortcite{tian2024visual}} & \cmark & \cmark & \xmark & 1.70\\
         \makecell{CART-d30} & \cmark & \cmark & \cmark & 1.57\\
         \hline
    \end{tabular}
    \caption{Ablation Study of CART}
    \label{tab:ablation1}
\end{table}

Figure \ref{fig:cartvvar_main} compares intermediate generations and self-attention maps for CART and VAR, the latter operating with multi-scale tokenization. VAR jointly refines global layout and local texture at each step, yielding entangled representations that hinder factor-wise control and scaling across resolutions. In contrast, CART first synthesizes a piecewise-smooth base capturing global structure, then incrementally adds detail factors, leading to an explicit hierarchy from structure to texture. This separation improves high-resolution synthesis via base upscaling with patch-wise detail prediction and enhances adaptability to target resolutions unseen during training. The tokenization order aligns with human perceptual organization, prioritizing coarse structures before fine details, and is reflected in progressively localized attention patterns in later steps 

\begin{figure}
\centering     

\includegraphics[width=0.49\textwidth]{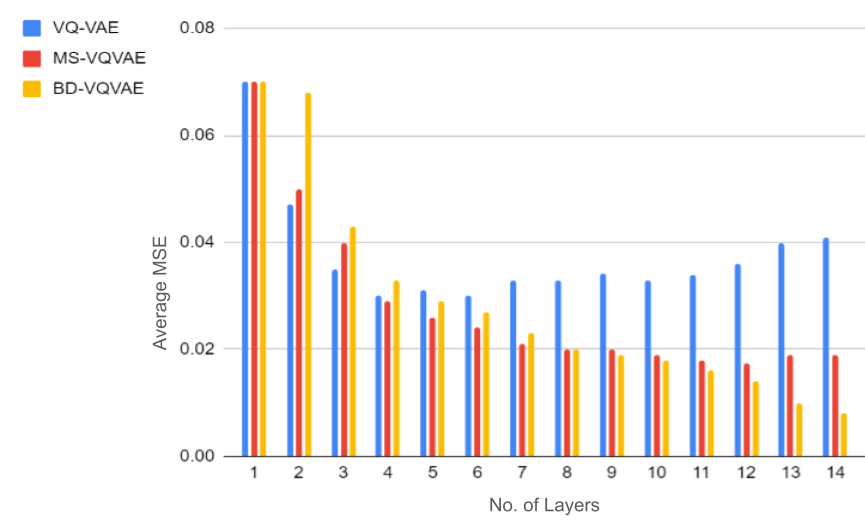} 
\caption{Reconstruction MSE of Vanilla VQ-VAE (blue), Multiscale VQ-VAE (red) and Base-Detail VQ-VAE (yellow)}
\label{fig:VQVAE_Compare_main}

\end{figure}

As shown in Figure \ref{fig:VQVAE_Compare_main}, the Base–Detail VQ-VAE attains substantially lower reconstruction error than both Vanilla VQ-VAE \cite{yu2021vector} and MS‑VQ-VAE \cite{tian2024visual}. Notably, while the MSE of Vanilla and MS‑VQ‑VAE increases at higher quantization depth, the proposed base–detail scheme continues to monotonically reduce reconstruction error, indicating more effective residual allocation and improved fidelity.


\begin{table}[]
    \centering
    \begin{tabular}{c|c}
    \hline
            Decomposition Order & FID \\
            \hline
             0 (Special case of VAR) & 1.70\\
             1  & 1.65\\
             2 & 1.62\\
             \hline
             3 & 1.57\\
             \hline
             4 & 1.60\\
    \end{tabular}
    \caption{Impact of decomposition order on CART model.}
    \label{tab:dec_order}
\end{table}







\section{Conclusion}


In this paper, we presented a novel auto-regressive framework with next-detail prediction and structured base-detail decomposition, enabling efficient, high-resolution image synthesis through iterative refinement. Our tokenization strategy of separately quantizing base and detail layers, preserves spatial integrity and enhances AR efficiency. Experiments show SOTA image generation and trainig-free extension to editing applications, surpassing limitations of next-token and next-scale approaches for accuracy and efficiency.

\bibliography{aaai2026}

@String(AAAI = {AAAI})

@article{ho2020denoising,
  title={Denoising diffusion probabilistic models},
  author={Ho, Jonathan and Jain, Ajay and Abbeel, Pieter},
  journal={Advances in neural information processing systems},
  volume={33},
  pages={6840--6851},
  year={2020}
}

@article{goodfellow2020generative,
  title={Generative adversarial networks},
  author={Goodfellow, Ian and Pouget-Abadie, Jean and Mirza, Mehdi and Xu, Bing and Warde-Farley, David and Ozair, Sherjil and Courville, Aaron and Bengio, Yoshua},
  journal={Communications of the ACM},
  volume={63},
  number={11},
  pages={139--144},
  year={2020},
  publisher={ACM New York, NY, USA}
}

@article{kingma2013auto,
  title={Auto-encoding variational bayes},
  author={Kingma, Diederik P},
  journal={arXiv preprint arXiv:1312.6114},
  year={2013}
}

@inproceedings{van2016pixel,
  title={Pixel recurrent neural networks},
  author={Van Den Oord, A{\"a}ron and Kalchbrenner, Nal and Kavukcuoglu, Koray},
  booktitle={International conference on machine learning},
  pages={1747--1756},
  year={2016},
  organization={PMLR}
}

@article{salimans2017pixelcnn++,
  title={Pixelcnn++: Improving the pixelcnn with discretized logistic mixture likelihood and other modifications},
  author={Salimans, Tim and Karpathy, Andrej and Chen, Xi and Kingma, Diederik P},
  journal={arXiv preprint arXiv:1701.05517},
  year={2017}
}

@inproceedings{gregor2015draw,
  title={Draw: A recurrent neural network for image generation},
  author={Gregor, Karol and Danihelka, Ivo and Graves, Alex and Rezende, Danilo and Wierstra, Daan},
  booktitle={International conference on machine learning},
  pages={1462--1471},
  year={2015},
  organization={PMLR}
}

@inproceedings{parmar2018image,
  title={Image transformer},
  author={Parmar, Niki and Vaswani, Ashish and Uszkoreit, Jakob and Kaiser, Lukasz and Shazeer, Noam and Ku, Alexander and Tran, Dustin},
  booktitle={International conference on machine learning},
  pages={4055--4064},
  year={2018},
  organization={PMLR}
}

@inproceedings{esser2021taming,
  title={Taming transformers for high-resolution image synthesis},
  author={Esser, Patrick and Rombach, Robin and Ommer, Bjorn and et al.},
  booktitle={Proceedings of the IEEE/CVF conference on computer vision and pattern recognition},
  pages={12873--12883},
  year={2021}
}

@inproceedings{ramesh2021zero,
  title={Zero-shot text-to-image generation},
  author={Ramesh, Aditya and Pavlov, Mikhail and Goh, Gabriel and Gray, Scott and Voss, Chelsea and Radford, Alec and Chen, Mark and Sutskever, Ilya},
  booktitle={International conference on machine learning},
  pages={8821--8831},
  year={2021},
  organization={Pmlr}
}

@article{tian2024visual,
  title={Visual autoregressive modeling: Scalable image generation via next-scale prediction},
  author={Tian, Keyu and Jiang, Yi and Yuan, Zehuan and Peng, Bingyue and Wang, Liwei},
  journal={Advances in neural information processing systems},
  volume={37},
  pages={84839--84865},
  year={2024}
}

@article{song2020denoising,
  title={Denoising diffusion implicit models},
  author={Song, Jiaming and Meng, Chenlin and Ermon, Stefano},
  journal={arXiv preprint arXiv:2010.02502},
  year={2020}
}

@article{brown2020language,
  title={Language models are few-shot learners},
  author={Brown, Tom B},
  journal={arXiv preprint arXiv:2005.14165},
  year={2020}
}

@article{radford2019language,
  title={Language models are unsupervised multitask learners},
  author={Radford, Alec and Wu, Jeffrey and Child, Rewon and Luan, David and Amodei, Dario and Sutskever, Ilya and others},
  journal={OpenAI blog},
  volume={1},
  number={8},
  pages={9},
  year={2019}
}

@article{vaswani2017attention,
  title={Attention is all you need},
  author={Vaswani, A},
  journal={Advances in Neural Information Processing Systems},
  year={2017}
}

@inproceedings{chen2020generative,
  title={Generative pretraining from pixels},
  author={Chen, Mark and Radford, Alec and Child, Rewon and Wu, Jeffrey and Jun, Heewoo and Luan, David and Sutskever, Ilya},
  booktitle={International conference on machine learning},
  pages={1691--1703},
  year={2020},
  organization={PMLR}
}

@article{van2017neural,
  title={Neural discrete representation learning},
  author={Van Den Oord, Aaron and Vinyals, Oriol and others},
  journal={Advances in neural information processing systems},
  volume={30},
  year={2017}
}

@inproceedings{zhang2018unreasonable,
  title={The unreasonable effectiveness of deep features as a perceptual metric},
  author={Zhang, Richard and Isola, Phillip and Efros, Alexei A and Shechtman, Eli and Wang, Oliver},
  booktitle={Proceedings of the IEEE conference on computer vision and pattern recognition},
  pages={586--595},
  year={2018}
}

@inproceedings{karras2019style,
  title={A style-based generator architecture for generative adversarial networks},
  author={Karras, Tero and Laine, Samuli and Aila, Timo},
  booktitle={Proceedings of the IEEE/CVF conference on computer vision and pattern recognition},
  pages={4401--4410},
  year={2019}
}

@article{mumford1989optimal,
  title={Optimal approximations by piecewise smooth functions and associated variational problems},
  author={Mumford, David Bryant and Shah, Jayant},
  journal={Communications on pure and applied mathematics},
  year={1989},
  publisher={Wiley-Blackwell}
}

@article{ambrosio1990approximation,
  title={Approximation of functional depending on jumps by elliptic functional via t-convergence},
  author={Ambrosio, Luigi and Tortorelli, Vincenzo Maria},
  journal={Communications on Pure and Applied Mathematics},
  volume={43},
  number={8},
  pages={999--1036},
  year={1990},
  publisher={Wiley Online Library}
}

@inproceedings{lee2022autoregressive,
  title={Autoregressive image generation using residual quantization},
  author={Lee, Doyup and Kim, Chiheon and Kim, Saehoon and Cho, Minsu and Han, Wook-Shin},
  booktitle={Proceedings of the IEEE/CVF Conference on Computer Vision and Pattern Recognition},
  pages={11523--11532},
  year={2022}
}

@article{kuznetsova2020open,
  title={The open images dataset v4: Unified image classification, object detection, and visual relationship detection at scale},
  author={Kuznetsova, Alina and Rom, Hassan and Alldrin, Neil and Uijlings, Jasper and Krasin, Ivan and Pont-Tuset, Jordi and Kamali, Shahab and Popov, Stefan and Malloci, Matteo and Kolesnikov, Alexander and others},
  journal={International journal of computer vision},
  volume={128},
  number={7},
  pages={1956--1981},
  year={2020},
  publisher={Springer}
}

@article{brock2018large,
  title={Large Scale GAN Training for High Fidelity Natural Image Synthesis},
  author={Brock, Andrew},
  journal={arXiv preprint arXiv:1809.11096},
  year={2018}
}

@inproceedings{kang2023scaling,
  title={Scaling up gans for text-to-image synthesis},
  author={Kang, Minguk and Zhu, Jun-Yan and Zhang, Richard and Park, Jaesik and Shechtman, Eli and Paris, Sylvain and Park, Taesung},
  booktitle={Proceedings of the IEEE/CVF Conference on Computer Vision and Pattern Recognition},
  pages={10124--10134},
  year={2023}
}

@inproceedings{sauer2022stylegan,
  title={Stylegan-xl: Scaling stylegan to large diverse datasets},
  author={Sauer, Axel and Schwarz, Katja and Geiger, Andreas and et al.},
  booktitle={ACM SIGGRAPH 2022 conference proceedings},
  pages={1--10},
  year={2022}
}

@article{dhariwal2021diffusion,
  title={Diffusion models beat gans on image synthesis},
  author={Dhariwal, Prafulla and Nichol, Alexander},
  journal={Advances in neural information processing systems},
  volume={34},
  pages={8780--8794},
  year={2021}
}

@article{huang2025spectralar,
  title={SpectralAR: Spectral Autoregressive Visual Generation},
  author={Huang, Yuanhui and Chen, Weiliang and Zheng, Wenzhao and Duan, Yueqi and Zhou, Jie and Lu, Jiwen},
  journal={arXiv preprint arXiv:2506.10962},
  year={2025}
}

@article{sun2024autoregressive,
  title={Autoregressive model beats diffusion: Llama for scalable image generation},
  author={Sun, Peize and Jiang, Yi and Chen, Shoufa and Zhang, Shilong and Peng, Bingyue and Luo, Ping and Yuan, Zehuan},
  journal={arXiv preprint arXiv:2406.06525},
  year={2024}
}

@inproceedings{wang2021real,
  title={Real-esrgan: Training real-world blind super-resolution with pure synthetic data},
  author={Wang, Xintao and Xie, Liangbin and Dong, Chao and Shan, Ying},
  booktitle={Proceedings of the IEEE/CVF international conference on computer vision},
  pages={1905--1914},
  year={2021}
}

@inproceedings{wang2024sinsr,
  title={Sinsr: diffusion-based image super-resolution in a single step},
  author={Wang, Yufei and Yang, Wenhan and Chen, Xinyuan and Wang, Yaohui and Guo, Lanqing and Chau, Lap-Pui and Liu, Ziwei and Qiao, Yu and Kot, Alex C and Wen, Bihan},
  booktitle={Proceedings of the IEEE/CVF conference on computer vision and pattern recognition},
  pages={25796--25805},
  year={2024}
}

@inproceedings{hatamizadeh2024diffit,
  title={Diffit: Diffusion vision transformers for image generation},
  author={Hatamizadeh, Ali and Song, Jiaming and Liu, Guilin and Kautz, Jan and Vahdat, Arash},
  booktitle={European Conference on Computer Vision},
  pages={37--55},
  year={2024},
  organization={Springer}
}

@article{ho2022cascaded,
  title={Cascaded diffusion models for high fidelity image generation},
  author={Ho, Jonathan and Saharia, Chitwan and Chan, William and Fleet, David J and Norouzi, Mohammad and Salimans, Tim},
  journal={Journal of Machine Learning Research},
  volume={23},
  number={47},
  pages={1--33},
  year={2022}
}

@inproceedings{rombach2022high,
  title={High-resolution image synthesis with latent diffusion models},
  author={Rombach, Robin and Blattmann, Andreas and Lorenz, Dominik and Esser, Patrick and Ommer, Bj{\"o}rn},
  booktitle={Proceedings of the IEEE/CVF conference on computer vision and pattern recognition},
  pages={10684--10695},
  year={2022}
}

@inproceedings{peebles2023scalable,
  title={Scalable diffusion models with transformers},
  author={Peebles, William and Xie, Saining},
  booktitle={Proceedings of the IEEE/CVF International Conference on Computer Vision},
  pages={4195--4205},
  year={2023}
}

@inproceedings{chang2022maskgit,
  title={Maskgit: Masked generative image transformer},
  author={Chang, Huiwen and Zhang, Han and Jiang, Lu and Liu, Ce and Freeman, William T},
  booktitle={Proceedings of the IEEE/CVF Conference on Computer Vision and Pattern Recognition},
  pages={11315--11325},
  year={2022}
}

@article{li2023self,
  title={Self-conditioned image generation via generating representations},
  author={Li, Tianhong and Katabi, Dina and He, Kaiming},
  journal={arXiv preprint arXiv:2312.03701},
  year={2023}
}

@article{razavi2019generating,
  title={Generating diverse high-fidelity images with vq-vae-2},
  author={Razavi, Ali and Van den Oord, Aaron and Vinyals, Oriol and et al.},
  journal={Advances in neural information processing systems},
  volume={32},
  year={2019}
}

@article{yu2021vector,
  title={Vector-quantized image modeling with improved vqgan},
  author={Yu, Jiahui and Li, Xin and Koh, Jing Yu and Zhang, Han and Pang, Ruoming and Qin, James and Ku, Alexander and Xu, Yuanzhong and Baldridge, Jason and Wu, Yonghui},
  journal={arXiv preprint arXiv:2110.04627},
  year={2021}
}

@inproceedings{lugmayr2022repaint,
  title={Repaint: Inpainting using denoising diffusion probabilistic models},
  author={Lugmayr, Andreas and Danelljan, Martin and Romero, Andres and Yu, Fisher and Timofte, Radu and Van Gool, Luc},
  booktitle={Proceedings of the IEEE/CVF conference on computer vision and pattern recognition},
  pages={11461--11471},
  year={2022}
}

@inproceedings{anciukevivcius2023renderdiffusion,
  title={Renderdiffusion: Image diffusion for 3d reconstruction, inpainting and generation},
  author={Anciukevi{\v{c}}ius, Titas and Xu, Zexiang and Fisher, Matthew and Henderson, Paul and Bilen, Hakan and Mitra, Niloy J and Guerrero, Paul},
  booktitle={Proceedings of the IEEE/CVF conference on computer vision and pattern recognition},
  pages={12608--12618},
  year={2023}
}

@inproceedings{corneanu2024latentpaint,
  title={Latentpaint: Image inpainting in latent space with diffusion models},
  author={Corneanu, Ciprian and Gadde, Raghudeep and Martinez, Aleix M},
  booktitle={Proceedings of the IEEE/CVF Winter Conference on Applications of Computer Vision},
  pages={4334--4343},
  year={2024}
}

@inproceedings{brooks2023instructpix2pix,
  title={Instructpix2pix: Learning to follow image editing instructions},
  author={Brooks, Tim and Holynski, Aleksander and Efros, Alexei A},
  booktitle={Proceedings of the IEEE/CVF Conference on Computer Vision and Pattern Recognition},
  pages={18392--18402},
  year={2023}
}

@article{li2022srdiff,
  title={Srdiff: Single image super-resolution with diffusion probabilistic models},
  author={Li, Haoying and Yang, Yifan and Chang, Meng and Chen, Shiqi and Feng, Huajun and Xu, Zhihai and Li, Qi and Chen, Yueting},
  journal={Neurocomputing},
  volume={479},
  pages={47--59},
  year={2022},
  publisher={Elsevier}
}

@inproceedings{zhu2023conditional,
  title={Conditional text image generation with diffusion models},
  author={Zhu, Yuanzhi and Li, Zhaohai and Wang, Tianwei and He, Mengchao and Yao, Cong},
  booktitle={Proceedings of the IEEE/CVF Conference on Computer Vision and Pattern Recognition},
  pages={14235--14245},
  year={2023}
}

@article{zhang2023text,
  title={Text-to-image diffusion models in generative ai: A survey},
  author={Zhang, Chenshuang and Zhang, Chaoning and Zhang, Mengchun and Kweon, In So},
  journal={arXiv preprint arXiv:2303.07909},
  year={2023}
}

@article{yue2024resshift,
  title={Resshift: Efficient diffusion model for image super-resolution by residual shifting},
  author={Yue, Zongsheng and Wang, Jianyi and Loy, Chen Change},
  journal={Advances in Neural Information Processing Systems},
  volume={36},
  year={2024}
}

@inproceedings{deng2009imagenet,
  title={Imagenet: A large-scale hierarchical image database},
  author={Deng, Jia and Dong, Wei and Socher, Richard and Li, Li-Jia and Li, Kai and Fei-Fei, Li},
  booktitle={2009 IEEE conference on computer vision and pattern recognition},
  pages={248--255},
  year={2009},
  organization={Ieee}
}

@inproceedings{wang2023exploring,
  title={Exploring clip for assessing the look and feel of images},
  author={Wang, Jianyi and Chan, Kelvin CK and Loy, Chen Change},
  booktitle={Proceedings of the AAAI conference on artificial intelligence},
  volume={37},
  number={2},
  pages={2555--2563},
  year={2023}
}

@misc{LargeDiT,
  author = {Zhang, Renrui},
  title = {Alpha-VLLM. Large-dit-imagenet.},
  year = {2024},
  howpublished = {\url{https://github.com/Alpha-VLLM/LLaMA2-Accessory/tree/main/Large-DiT-ImageNet}},
  note = {Accessed: 2025-07-26}
}

@article{mirza2014conditional,
  title={Conditional generative adversarial nets},
  author={Mirza, Mehdi},
  journal={arXiv preprint arXiv:1411.1784},
  year={2014}
}

@inproceedings{shao2020controlvae,
  title={Controlvae: Controllable variational autoencoder},
  author={Shao, Huajie and Yao, Shuochao and Sun, Dachun and Zhang, Aston and Liu, Shengzhong and Liu, Dongxin and Wang, Jun and Abdelzaher, Tarek},
  booktitle={International conference on machine learning},
  pages={8655--8664},
  year={2020},
  organization={PMLR}
}

@inproceedings{yang2023uni,
  title={Uni-paint: A unified framework for multimodal image inpainting with pretrained diffusion model},
  author={Yang, Shiyuan and Chen, Xiaodong and Liao, Jing},
  booktitle={Proceedings of the 31st ACM International Conference on Multimedia},
  pages={3190--3199},
  year={2023}
}

@article{careaga2023intrinsic,
  title={Intrinsic image decomposition via ordinal shading},
  author={Careaga, Chris and Aksoy, Ya{\u{g}}{\i}z},
  journal={ACM Transactions on Graphics},
  volume={43},
  number={1},
  pages={1--24},
  year={2023},
  publisher={ACM New York, NY, USA}
}

@inproceedings{kawar2023imagic,
  title={Imagic: Text-based real image editing with diffusion models},
  author={Kawar, Bahjat and Zada, Shiran and Lang, Oran and Tov, Omer and Chang, Huiwen and Dekel, Tali and Mosseri, Inbar and Irani, Michal},
  booktitle={Proceedings of the IEEE/CVF Conference on Computer Vision and Pattern Recognition},
  pages={6007--6017},
  year={2023}
}

@inproceedings{zhou2023sparsefusion,
  title={Sparsefusion: Distilling view-conditioned diffusion for 3d reconstruction},
  author={Zhou, Zhizhuo and Tulsiani, Shubham},
  booktitle={Proceedings of the IEEE/CVF Conference on Computer Vision and Pattern Recognition},
  pages={12588--12597},
  year={2023}
}

@article{nash2021generating,
  title={Generating images with sparse representations},
  author={Nash, Charlie and et al.},
  journal={arXiv preprint arXiv:2103.03841},
  year={2021}
}

@inproceedings{yu2021wavefill,
  title={Wavefill: A wavelet-based generation network for image inpainting},
  author={Yu, Yingchen and Zhan, Fangneng and Lu, Shijian and Pan, Jianxiong and Ma, Feiying and Xie, Xuansong and Miao, Chunyan},
  booktitle={Proceedings of the IEEE/CVF international conference on computer vision},
  pages={14114--14123},
  year={2021}
}

@inproceedings{saini2024specularity,
  title={Specularity factorization for low-light enhancement},
  author={Saini, Saurabh and Narayanan, PJ},
  booktitle={Proceedings of the IEEE/CVF Conference on Computer Vision and Pattern Recognition},
  pages={1--12},
  year={2024}
}

@article{tominaga1994dichromatic,
  title={Dichromatic reflection models for a variety of materials},
  author={Tominaga, Shoji},
  journal={Color Research \& Application},
  volume={19},
  number={4},
  pages={277--285},
  year={1994},
  publisher={Wiley Online Library}
}

@article{bala2024galaxyedit,
  title={GalaxyEdit: Large-Scale Image Editing Dataset with Enhanced Diffusion Adapter},
  author={Bala, Aniruddha and Jaiswal, Rohan and Rashid, Loay and Roheda, Siddharth},
  journal={arXiv preprint arXiv:2411.13794},
  year={2024}
}
\clearpage
\setcounter{page}{1}


\twocolumn[{%
\renewcommand\twocolumn[1][]{#1}%
\justify{
\section{Supplementary: CART: Compositional Auto Regressive Transformer for Image Generation}
\text{ }
\text{ }
\section{Metric Definitions}
Here we formally define the metrics utilized to evaluate our approach against other SOTA techniques:
\begin{itemize}
    \item \textbf{Fréchet Inception Distance} (FID $\downarrow$): FID is a quantitative metric used to assess the quality of images generated by probabilistic models. FID measures the similarity between the distributions of features extracted from real and generated images using the Inception v3 network. Let $\mathcal{X}_{\text{real}} = \{x_i\}_{i=1}^N$ be a set of real images, $\mathcal{X}_{\text{gen}} = \{\tilde{x}_i\}_{i=1}^M$ be a set of generated images, and $f(\cdot)$ denote the feature extraction function of the Inception v3 network. Define the empirical means and covariances of the extracted features as $\mu_r$ and $\Sigma_r$ for $X_{real}$ and $\mu_g$ and $\Sigma_g$ for $X_{gen}$. The FID between the real and generated image sets is then defined as,
    \begin{equation}
        \mathrm{FID}^2 = \|\mu_r - \mu_g\|_2^2 + \mathrm{Tr}\left(\Sigma_r + \Sigma_g - 2 \left(\Sigma_r \Sigma_g \right)^{\frac{1}{2}} \right)
    \end{equation}
    where $\|\cdot\|_2$ is the Euclidean norm, $\mathrm{Tr}(\cdot)$ denotes the matrix trace, and $\left(\Sigma_r \Sigma_g \right)^{\frac{1}{2}}$ is the unique positive semi-definite square root of the matrix product $\Sigma_r \Sigma_g$.

    A lower FID indicates a closer match between generated and real image distributions in the Inception feature space, thus reflecting better generative model quality.


    \item \textbf{Inception Score} (IS $\uparrow$): The Inception Score (IS) is a widely used metric for evaluating the quality and diversity of images generated by a generative model. It uses a pretrained image classification network (typically Inception v3) to assess the generated samples. Let $\mathcal{X}_{\text{gen}} = \{x_i\}_{i=1}^N$ be a set of generated images, $p(y|x)$ denote the conditional label distribution predicted by the Inception v3 model for image $x$, where $y$ indexes the class labels and $p(y) = \frac{1}{N} \sum_{i=1}^N p(y|x_i)$ be the marginal distribution of predicted labels over all generated images. The Inception Score is defined as, 

    \begin{equation}
        \mathrm{IS} = \exp \left( \mathbb{E}_{x \sim \mathcal{X}_{\text{gen}}} \left[ D_{\mathrm{KL}} \big(p(y|x) \| p(y)\big) \right] \right),
    \end{equation}
    where $D_{\mathrm{KL}}(\cdot \| \cdot)$ is the Kullback–Leibler divergence, and $\mathbb{E}_{x \sim \mathcal{X}_{\text{gen}}}[\cdot]$ denotes expectation over generated images.

    A higher Inception Score reflects that generated images are both \textit{highly classifiable} (each image has low-entropy, confident predictions) and \textit{diverse} across classes (high entropy for the marginal distribution), indicating good generative model performance.
    
    \item \textbf{CLIP Image Quality Assesment Score} (CLIPIQA $\uparrow$) is a no-reference image quality metric that leverages the joint multimodal embedding space of the CLIP (Contrastive Language-Image Pretraining) model for perceptual image quality assessment. It measures image quality by comparing the similarity between an image and carefully designed quality-related textual prompts within the CLIP embedding space. Let $I$ be the input image, $T_1$ and $T_2$ be a pair of antonym text prompts (e.g., ``Good photo.'' and ``Bad photo.''), and $x = E_{\mathrm{img}}(I)$ be the normalized CLIP image embedding of $I$. Further, let $t_i = E_{\mathrm{text}}(T_i)$ be the normalized CLIP text embedding of prompt $T_i$, for $i \in \{1,2\}$. The cosine similarity between $x$ and $t_i$ is calculated as $s_i = x^\top t_i$. Finally, the CLIPIQA score for image $I$ is computed as, 
    \begin{equation}
        \bar{s} = \frac{\exp(s_1)}{\exp(s_1) + \exp(s_2)}.
    \end{equation}
    A higher CLIPIQA score $\bar{s}$ indicates that the image $I$ aligns more closely with the positive attribute in the CLIP embedding space, reflecting better perceptual quality, aesthetics, or specified abstract properties.

    \item \textbf{Peak Signal to Noise Ratio} (PSNR $\uparrow$): PSNR is a widely used metric to evaluate the quality of a reconstructed image relative to a reference original. It quantifies the ratio between the maximum possible pixel value and the mean squared error (MSE) introduced by reconstruction. Let $I$ be the original image and $I_R$ be the reconstructed image, both of size $m \times n$. The \textit{mean squared error} is defined as, $\mathrm{MSE} = \frac{1}{mn} \sum_{i=0}^{m-1} \sum_{j=0}^{n-1} \bigl(I(i,j) - I_R(i,j)\bigr)^2.$ The PSNR, expressed in decibels (dB), is calculated as,
    \begin{equation}
        \mathrm{PSNR} = 20 \cdot \log_{10} \left( \frac{\mathrm{MAX}_I}{\sqrt{\mathrm{MSE}}} \right).
    \end{equation}
    A higher PSNR indicates better reconstruction quality with less distortion.
    \item \textbf{Structural Similarity} (SSIM $\uparrow$): SSIM is a perceptual metric used to quantify the similarity between two images, typically a reference image $x$ and a test image $y$. SSIM measures image quality by considering changes in structural information while incorporating luminance and contrast masking, which aligns well with human visual perception. SSIM considers luminance similarity, $l(x,y) = \frac{2\mu_x \mu_y + C_1}{\mu_x^2 + \mu_y^2 + C_1}$, contrast similarity, $c(x,y) = \frac{2\sigma_x \sigma_y + C_2}{\sigma_x^2 + \sigma_y^2 + C_2}$, and strucural similarity $s(x,y) = \frac{\sigma_{xy} + C_3}{\sigma_x \sigma_y + C_3}$. Where $\mu_x, \mu_y$ are the mean intensities, $\sigma_x, \sigma_y$ are standard deviations, and $\sigma_{xy}$ is the covariance of $x$ and $y$. Constants $C_1, C_2, C_3$ stabilize divisions to avoid instability when denominators are close to zero. The SSIM index is computed as a multiplicative combination of these components, $\text{SSIM}(x,y) = [l(x,y)]^{\alpha} \cdot [c(x,y)]^{\beta} \cdot [s(x,y)]^{\gamma}$. The SSIM value ranges from 0 to 1, with 1 indicating perfect structural similarity.
\end{itemize}
}
}]

\begin{figure*}
\centering     
\includegraphics[width=\textwidth]{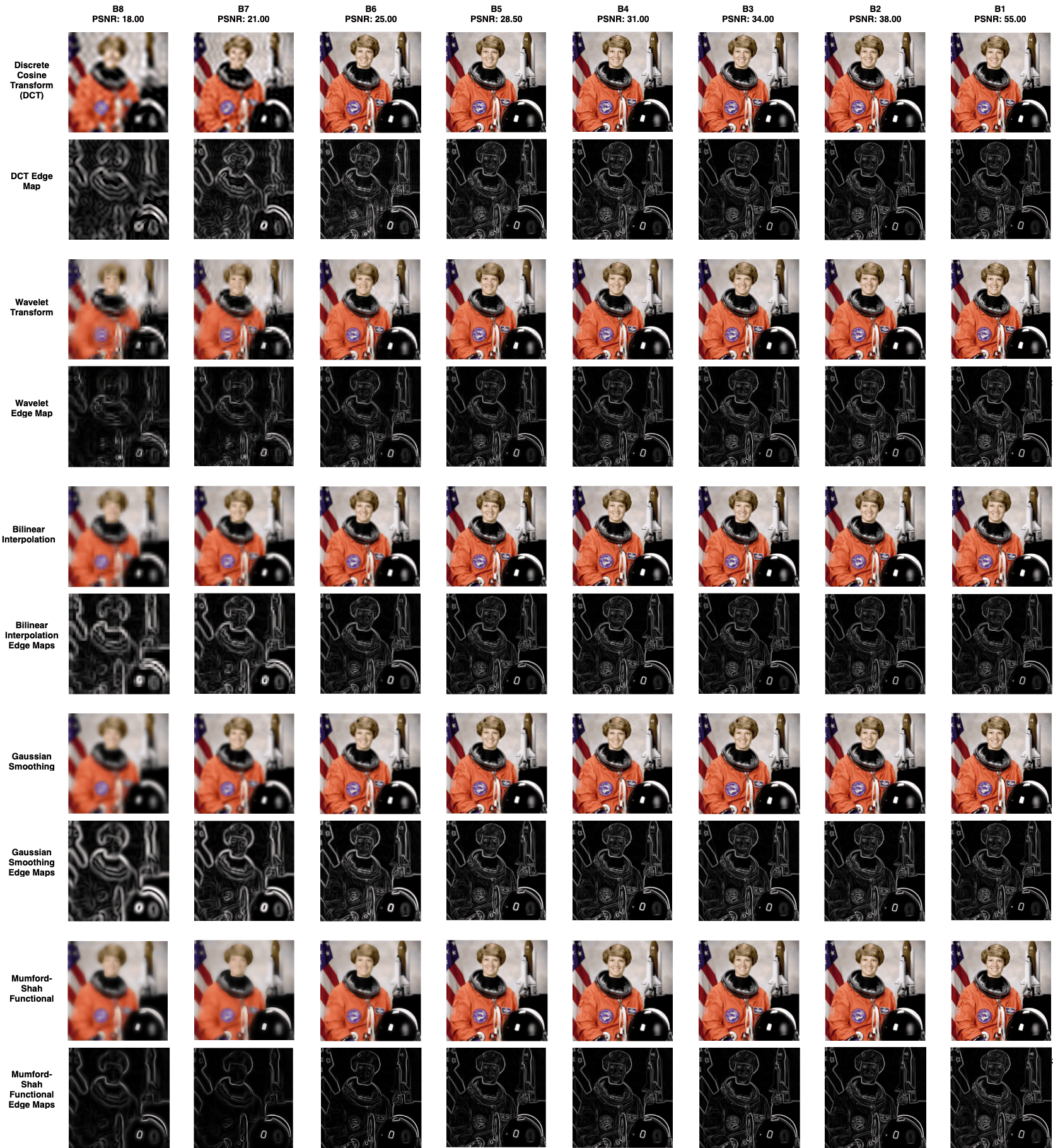}
\caption{Hierarchical smoothing using five different techniques. Zoom in recomended to observe fine details.}
\label{fig:hierarchical_smoothing}
\end{figure*}

\begin{figure*}
\centering     
\subfigure[Ringing Artifact Score]{\label{fig:ringing}\includegraphics[width=0.49\textwidth]{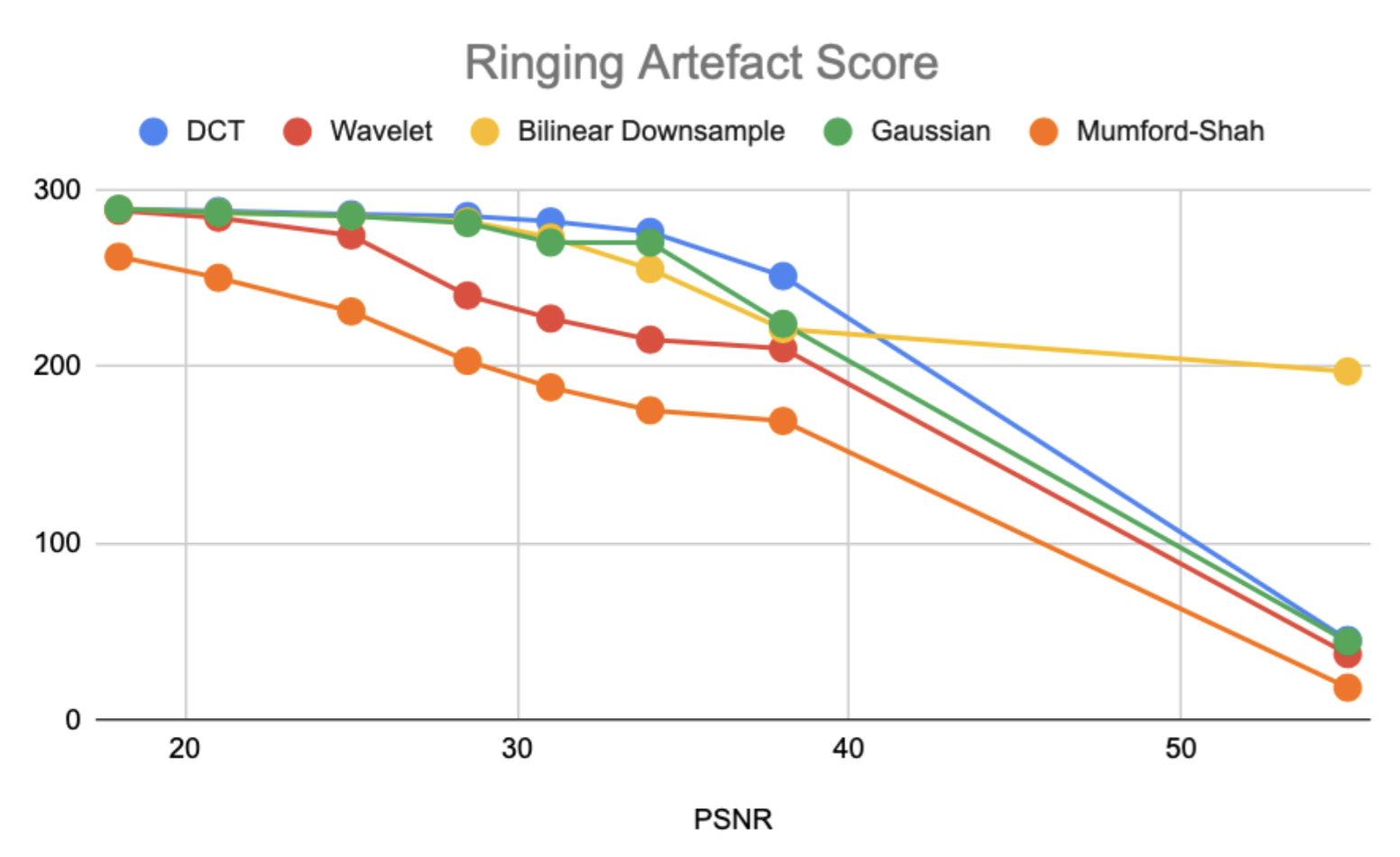}}
\subfigure[Gradient Stability Score]{\label{fig:grad}\includegraphics[width=0.49\textwidth]{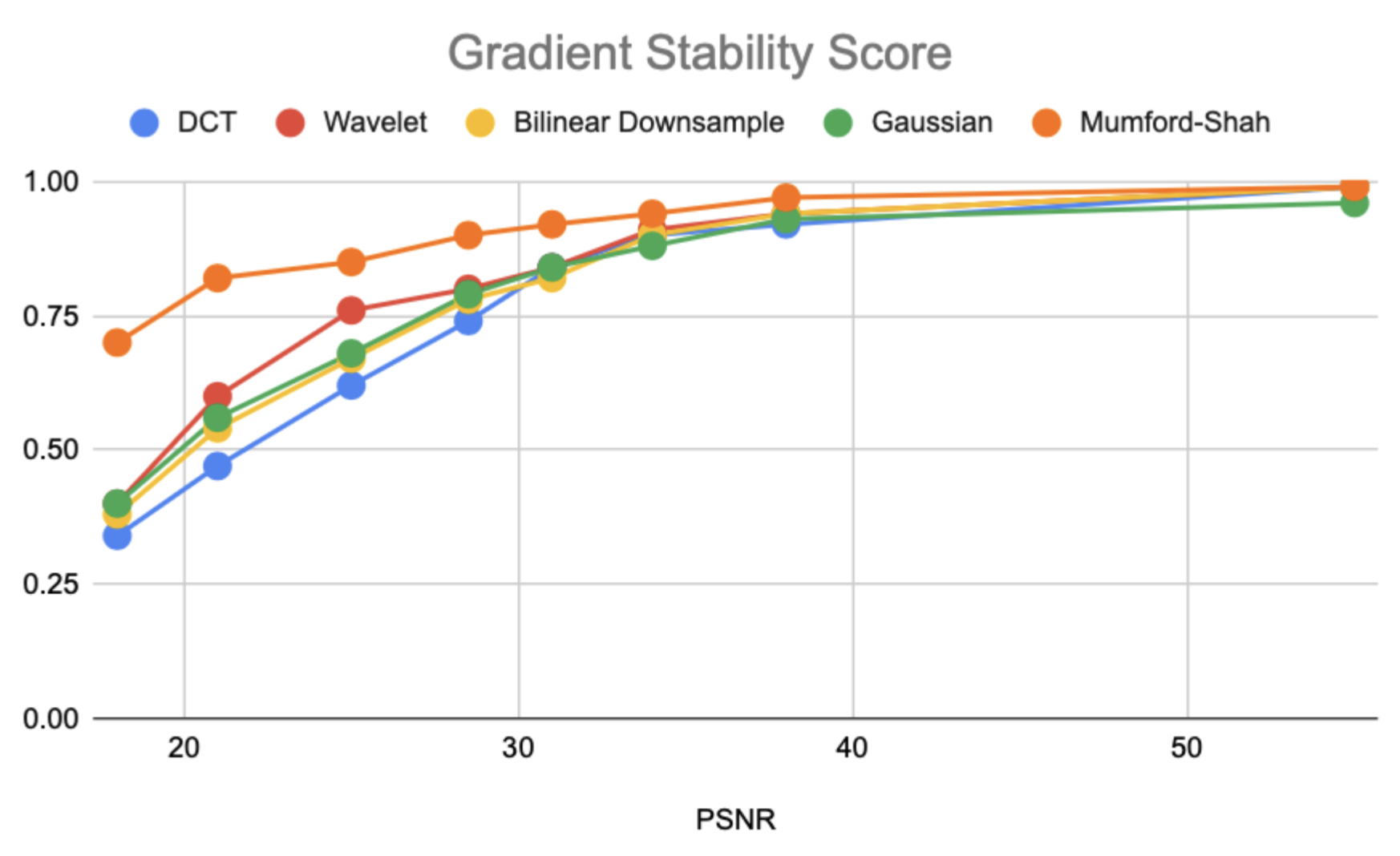}}
\caption{Ringing and Gradient Stability scores across the 8 levels depicted in Figure \ref{fig:hierarchical_smoothing}}
\label{fig:ms_graphs}
\end{figure*}

\begin{figure*}
\centering     

\includegraphics[width=0.9\textwidth]{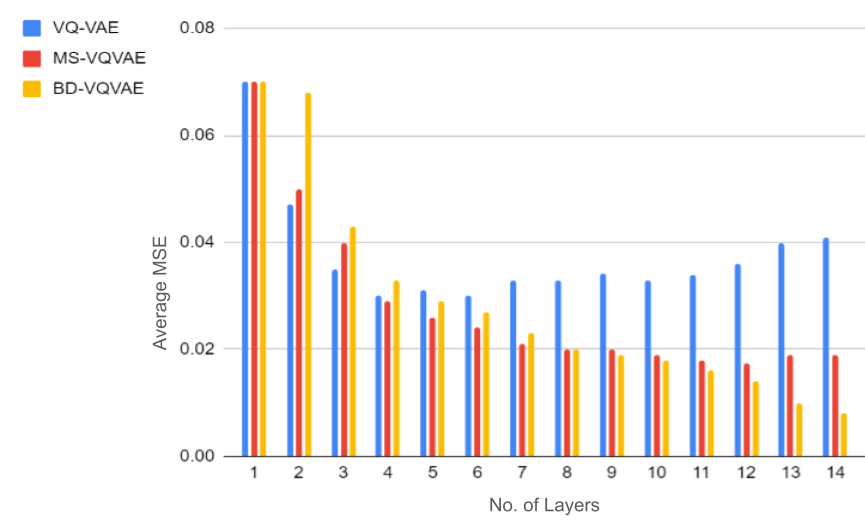} 
\caption{Reconstruction MSE of Vanilla VQ-VAE (blue), Multiscale VQ-VAE (red) and Base-Detail VQ-VAE (yellow)}
\label{fig:VQVAE_Compare}

\end{figure*}

\twocolumn[{%
\renewcommand\twocolumn[1][]{#1}%
\justify{
\section{Analysis of Smoothing Functions for Base-Detail Decomposition}
In this section, we present a comparative analysis of several candidate smoothing functions for base-detail decomposition as defined in Equation \ref{eq:nth_order_decomp}. Figure \ref{fig:hierarchical_smoothing} illustrates hierarchical smoothing results obtained via five distinct smoothing techniques along with their corresponding edge maps. For each smoothing level $B_k$, hyper-parameters are calibrated such that all methods achieve equivalent peak signal-to-noise ratio (PSNR) values, ensuring a fair and consistent comparison. Figure \ref{fig:ms_graphs} quantifies ringing artifacts and gradient stability across the eight smoothing levels visualized in Figure \ref{fig:hierarchical_smoothing}. 
As evident from the comparisons in Figure \ref{fig:hierarchical_smoothing}, Mumford--Shah smoothing yields a piecewise-smooth solution, which provides several critical advantages in the training of CART:

\begin{itemize}
    \item \textbf{Explicit Structural-Textural Disentanglement:} Mumford--Shah smoothing facilitates a clear separation between global structures and local details. This is verifiable in the edge maps progressing from $B_8 \to B_1$ in Figure \ref{fig:hierarchical_smoothing}, where early edge maps predominantly capture major structural edges while finer details progressively emerge at later stages. Such hierarchical decomposition aligns well with the ``next-detail'' prediction paradigm central to CART. In contrast, alternative methods like discrete cosine transform (DCT), bilinear downsampling, and Gaussian blurring exhibit more uniform smoothing, resulting in entanglement of global and local features at each smoothing level.
    
   
    \item \textbf{Artifact Suppression:} The Mumford--Shah functional's enforcement of piecewise-smoothness effectively mitigates ringing artifacts, which are prominent in DCT and wavelet-based smoothing due to the Gibbs phenomenon. Visual evidence of these ringing effects is presented in Figure \ref{fig:hierarchical_smoothing}, while their quantitative assessment in Figure \ref{fig:ringing} demonstrates Mumford--Shah smoothing's superiority in minimizing such artifacts. This clean base layer decomposition enhances training stability and supports training-free high-resolution image generation and super-resolution tasks.
    

    \item \textbf{Gradient Stability:} Owing to its inherent piecewise-smooth formulation, Mumford--Shah smoothing achieves superior gradient stability compared to competing smoothing approaches (see Figure \ref{fig:grad}), enabling more precise and accelerated model convergence and resulting in highly detailed, artifact-free image synthesis.
    
    
    \item \textbf{Reduced Reconstruction Error:} Empirical evaluation reveals that the Mumford--Shah based Base-Detail VQ-VAE outperforms multi-scale VQ-VAE in reconstruction fidelity, particularly at increased residual quantization depths (see Figure \ref{fig:VQVAE_Compare}).
    
\end{itemize}
\subsubsection{Computation of Ringing Artifact Score:} The ringing artifact score quantifies oscillatory distortions near edges in the smoothed base layer. A higher score corresponds to more pronounced ringing. For a given smoothed base layer $I_s$ and original image $I$, the computation proceeds as follows:
\begin{enumerate}
    \item Detect strong edges (top 20\%) in original image using sobel filter, $I_E$. 
    \item  Dilate obtained edge mask to include surrounding region which is candidate region for ringing, $I_D$.
    \item Calculate difference between Original and smooth base, $I_{Diff} = I - I_s$.
    \item Compute three measures of ringing: 
    \begin{enumerate}
        \item Apply High-Pass filter to the difference image in order to detect high-frequency oscillations and compute standard deviation around relevant edges, $L_{HP} = stddev(HP(L_{Diff}) [I_D])$
        \item Apply FFT to convert difference imge to frequency domain, and compute difference in frequency domain for mid to high range frequencies, $L_{FFT} = mean(FFT(I_{Diff})[Freq_Mask])$
        \item Calculate variance around edges in neighboring pixels, $L_{var} = mean(var(I_{Diff})[I_D])$
    \end{enumerate}
    \item Compute final ringing score, $L_{ringing} = \alpha*L_{HP} + \beta*L_{FFT} + \gamma * L_{var}$.
\end{enumerate}

\subsubsection{Computation of Gradient Stability Scores: } The gradient stability score assesses the preservation of image gradients in the smoothed base layer. A higher value indicates better gradient preservation. Given the original image $I$ and smoothed base $I_s$:
\begin{enumerate}
    \item Calculate the gradients of original image, $I_G = grad(I)$ and the base layer, $I_{s_G} = grad(I_s)$.
    \item Calculate the gradient stability score as the Pearson correlation coefficient between $I_G$ and $I_{s_G}$, $L_{Gradient} = Corr(I_G, I_{s_G})$.
    
\end{enumerate}

This rigorous evaluation substantiates Mumford--Shah smoothing as the preferred approach for base-detail decomposition within CART, balancing structural clarity, artifact mitigation, and gradient stability, thereby enhancing downstream autoregressive modeling and reconstruction fidelity.



}
}]

\begin{figure*}
\centering
    \includegraphics[width=0.95\textwidth]{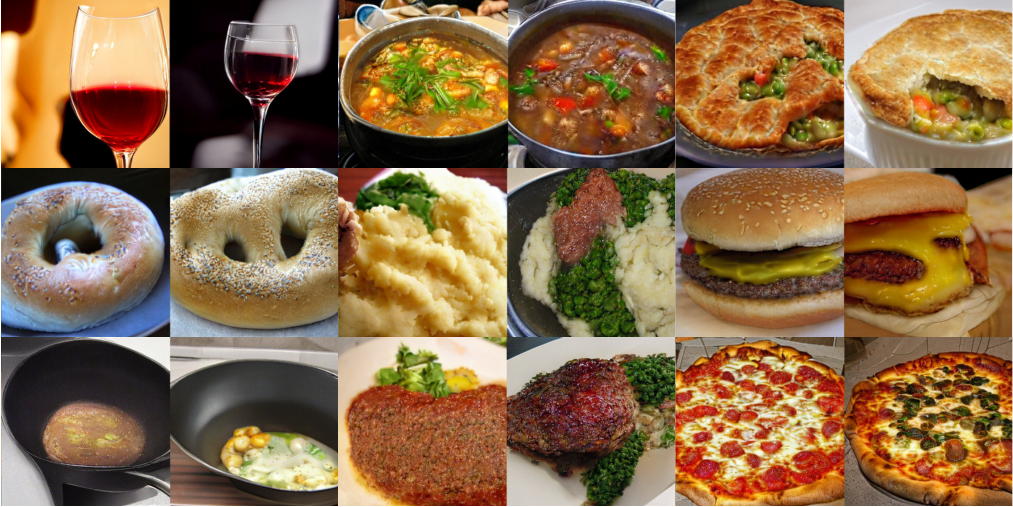}
    \caption{Generated Samples using CART}
    \label{fig:CART_Results7}

\end{figure*}

\twocolumn[{%
\renewcommand\twocolumn[1][]{#1}%

\section{Extended Results}
\label{sec:extended_results}
Figures \ref{fig:CART_Results2} - \ref{fig:CART_Results7} cover some generations from the CART model. These generations show the diversity of the CART model and how it generalizes over many classes while producing samples with enhanced details. 

\begin{center}
    \centering
    \captionsetup{type=figure}
    \includegraphics[width=.9\textwidth]{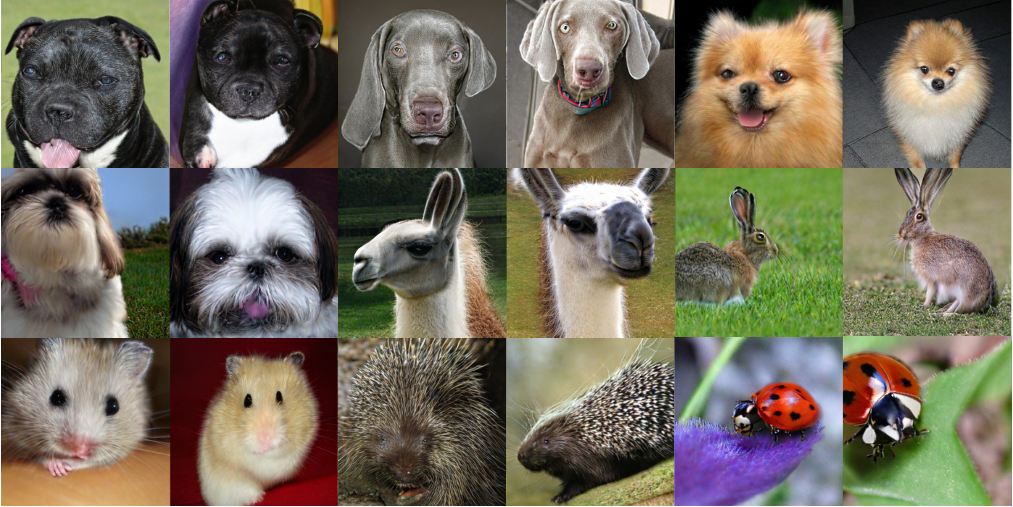}
    \captionof{figure}{Generated Samples using CART}
    \label{fig:CART_Results2}

    \includegraphics[width=0.9\textwidth]{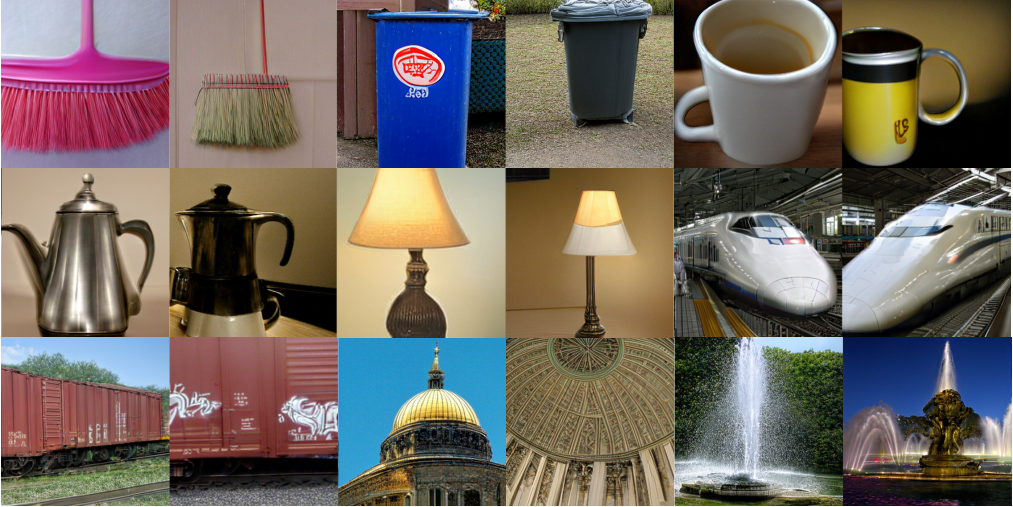}
    \caption{Generated Samples using CART}
    \label{fig:CART_Results3}
    
\end{center}%
}]



\begin{figure*}
\centering
    \includegraphics[width=0.95\textwidth]{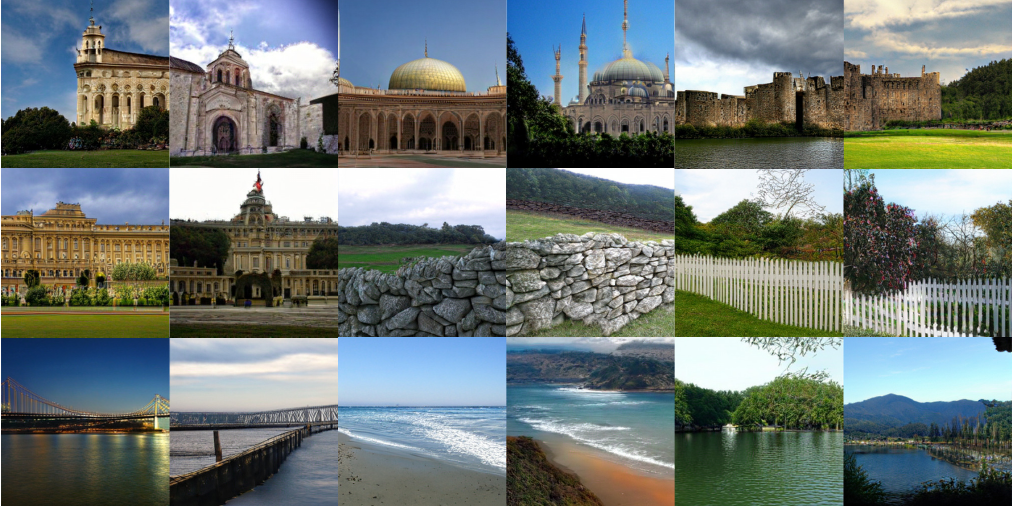}
    \caption{Generated Samples using CART}
    \label{fig:CART_Results4}

\end{figure*}

\begin{figure*}
\centering
    \includegraphics[width=0.95\textwidth]{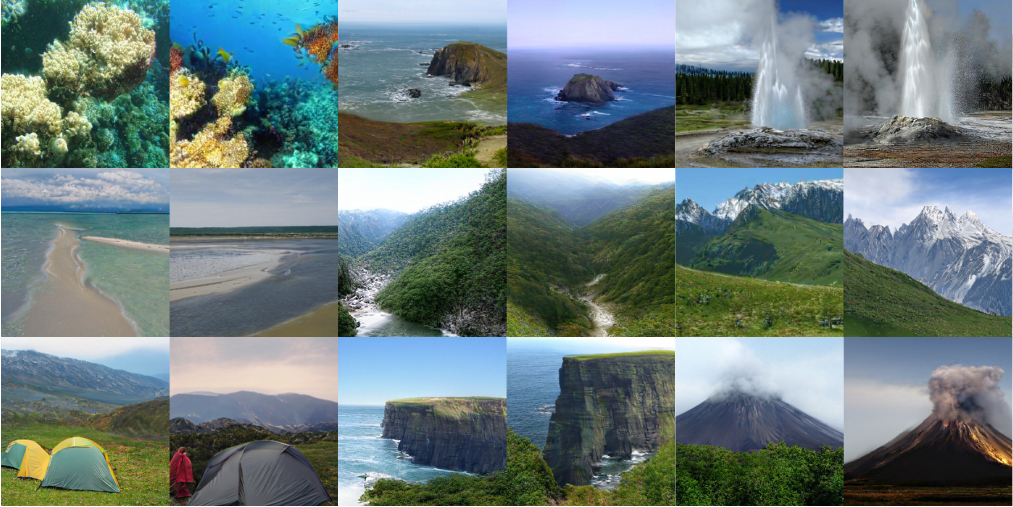}
    \caption{Generated Samples using CART}
    \label{fig:CART_Results5}

\end{figure*}

\begin{figure*}
\centering
    \includegraphics[width=0.95\textwidth]{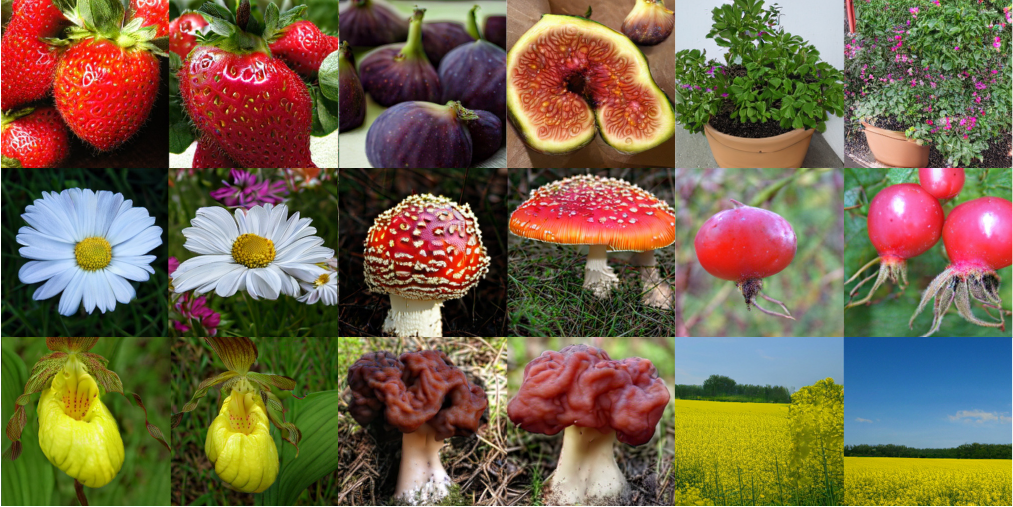}
    \caption{Generated Samples using CART}
    \label{fig:CART_Results6}

\end{figure*}

\begin{figure*}
\centering
    \includegraphics[width=0.9\textwidth]{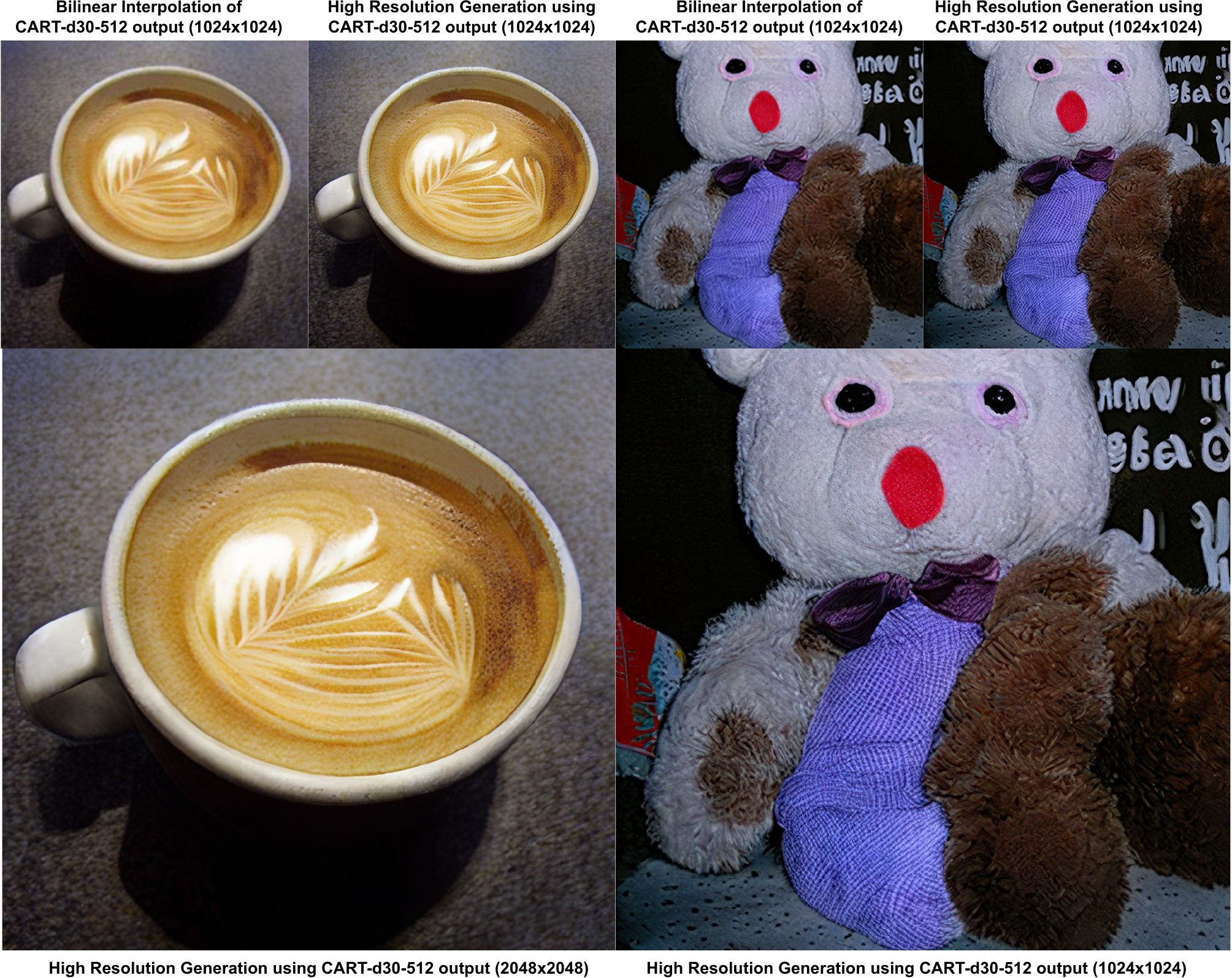}
    \caption{Generating high resolution images at $1024 \times 1024$ and $2048 \times 2048$ using CART-d30-512 model trained on $512 \times 512$ resolution. Left: Espresso; Right: Teddy}
    \label{fig:high_res_2}

\end{figure*}

\twocolumn[{%
\renewcommand\twocolumn[1][]{#1}%
\justify{
\section{High Resolution Image Generation}
\label{sec:high_res_ext}
The approach discussed in section \ref{sec:high_res} can be used to generate high-resolution images without any retraining of the base model as depicted in Figure \ref{fig:high_res_gen}. In Figures \ref{fig:high_res_1} and \ref{fig:high_res_2} we depict the generation of images at resolutions $1024 \times 1024$ and $2048 \times 2048$ using the CART-d30 model trained at $512 \times 512$ model. Note that we are capable of maintaining the same image content as the low resolution image as we re-use the base image generated at the lower resolution and only introduce patch-wise details at the desired higher-resolution. 
}

\begin{center}
    \centering
    \captionsetup{type=figure}
    \includegraphics[width=.9\textwidth]{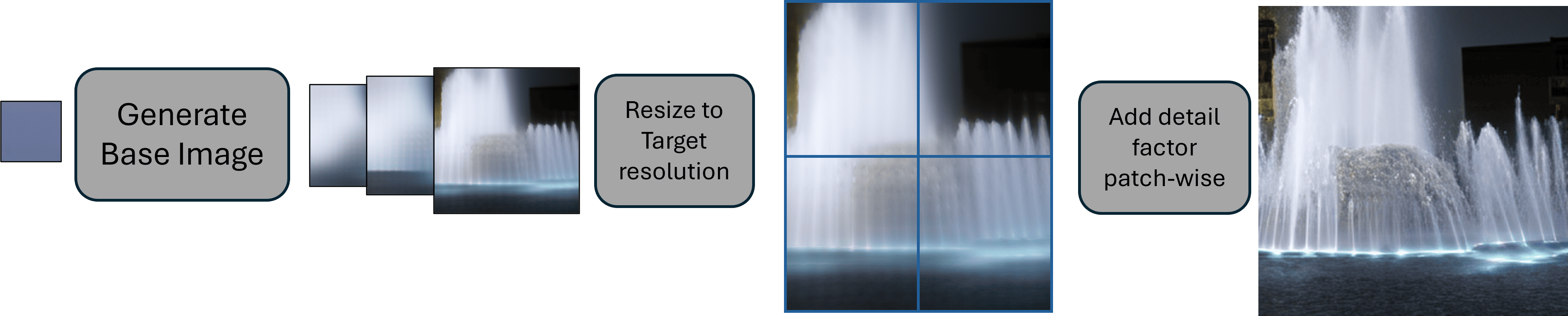}
    \captionof{figure}{Generation of high resolution images using CART models trained on low-resolution images}
    \label{fig:high_res_gen}

    \includegraphics[width=0.9\textwidth]{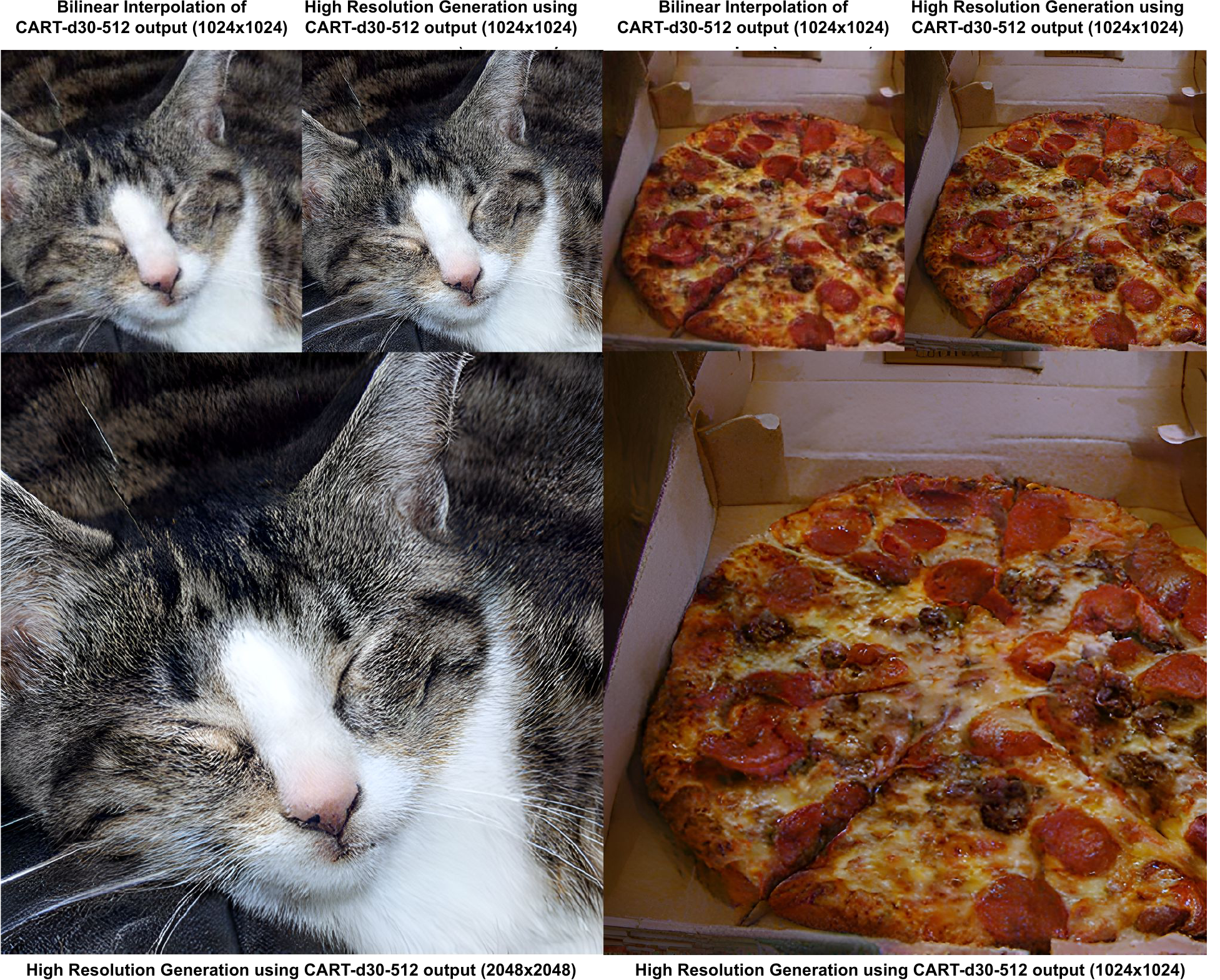}
    \caption{Generating high resolution images at $1024 \times 1024$ and $2048 \times 2048$ using CART-d30-512 model trained on $512 \times 512$ resolution. Left: Cat; Right: Pizza}
    \label{fig:high_res_1}
    
\end{center}%

}]

\begin{figure*}
\centering     
\includegraphics[width=\textwidth]{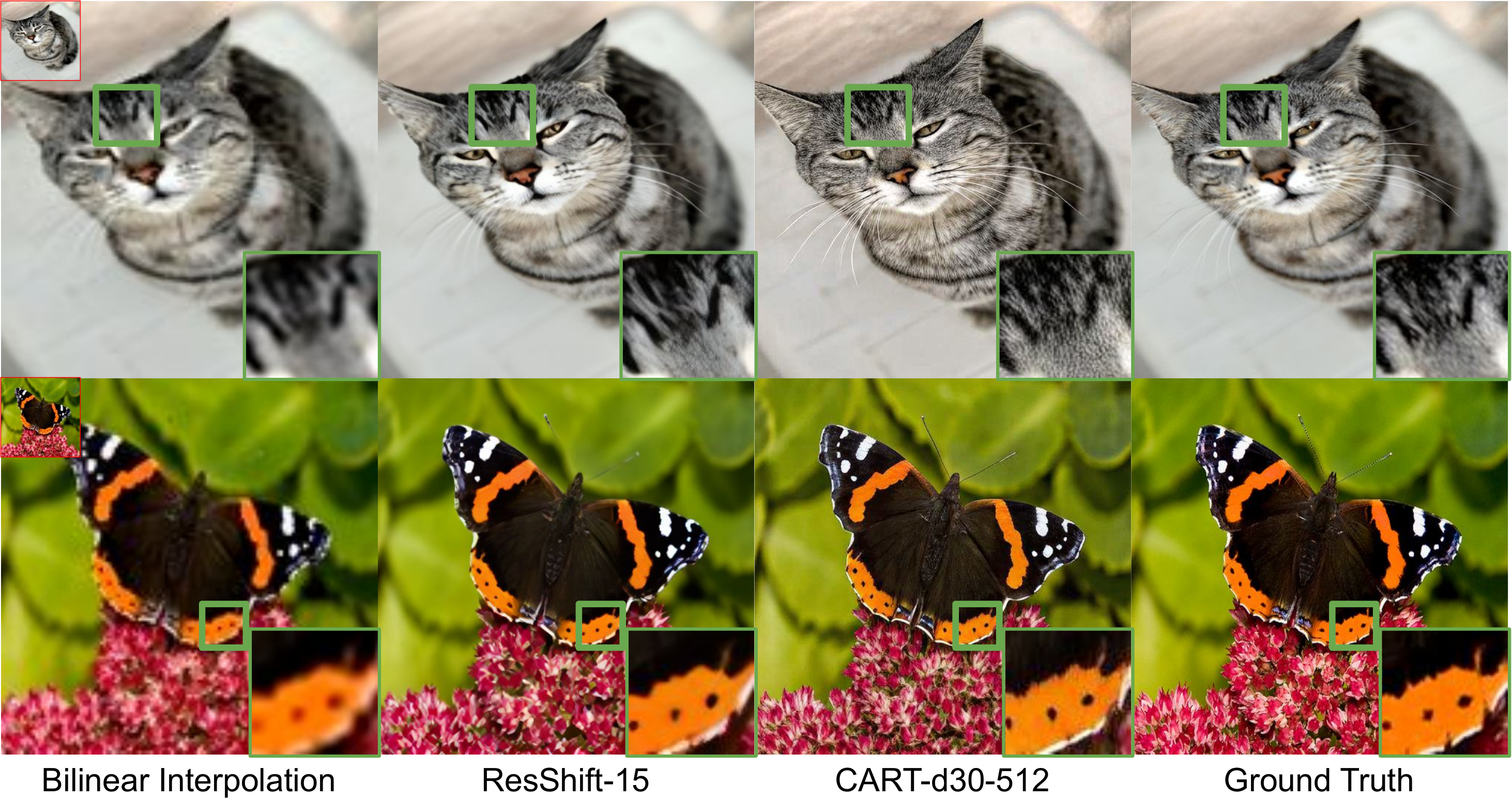}
\caption{Comparison of CART for super-resolution with ResShift. Zoom-in recommended to observe finer details.}
\label{fig:sr_supp5}
\end{figure*}

\twocolumn[{
\renewcommand\twocolumn[1][]{#1}%
\justify{
\section{Extended Results for Image Super-Resolution Using CART}
Figures \ref{fig:sr_supp1} through \ref{fig:sr_supp5} present the results of image super-resolution achieved using the CART-d30-512 model. The visual comparisons clearly demonstrate that CART-d30-512 produces images with enhanced textures and superior perceptual quality compared to those generated by the diffusion-based ResShift-15 method.
\begin{center}
    \centering
    \captionsetup{type=figure}
    \includegraphics[width=\textwidth]{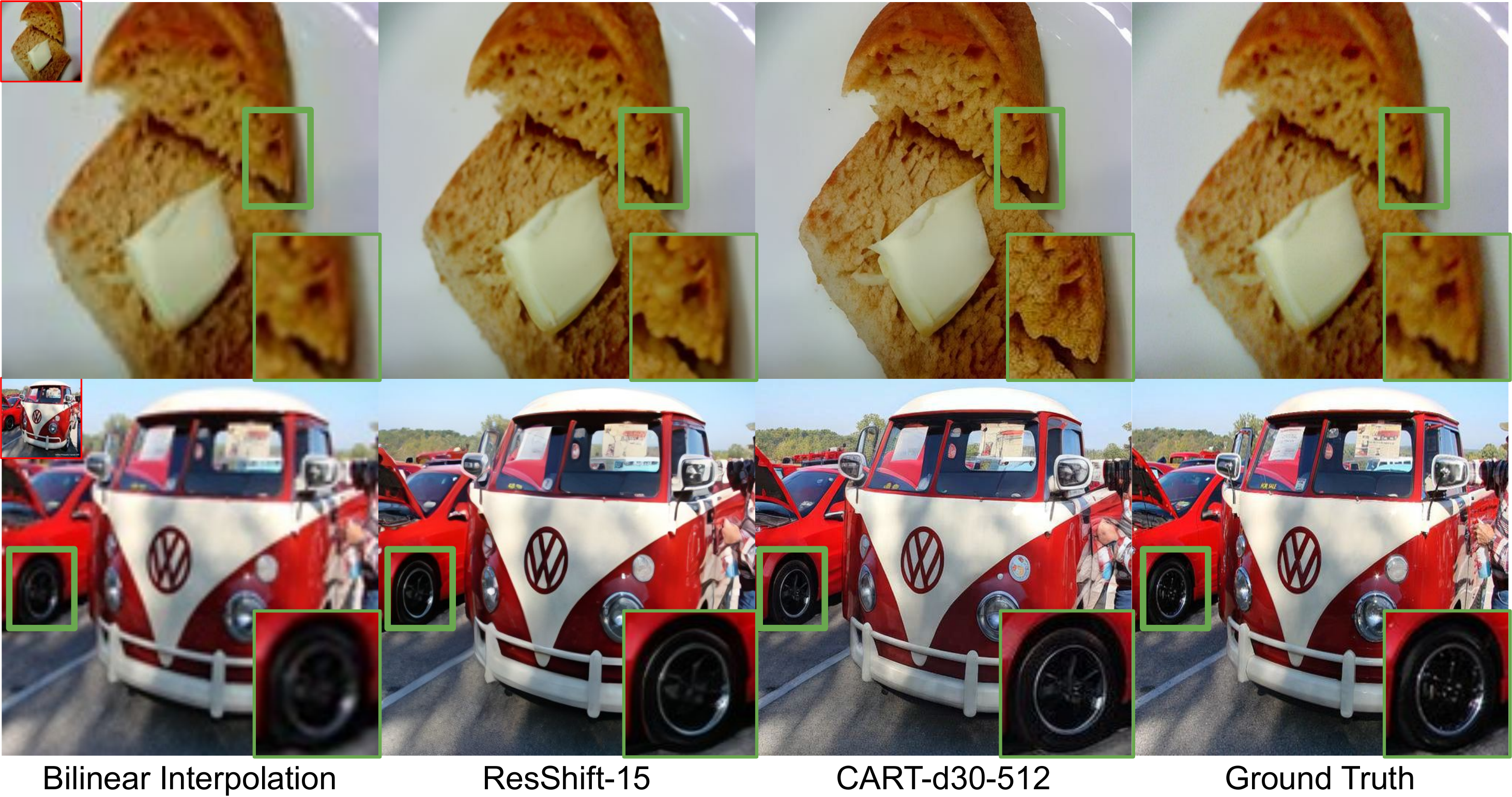}
    \captionof{figure}{Comparison of CART for super-resolution with ResShift. Zoom-in recommended to observe finer details. }
    \label{fig:sr_supp1}
\end{center}

\begin{center}
    \centering
    \captionsetup{type=figure}
    \includegraphics[width=\textwidth]{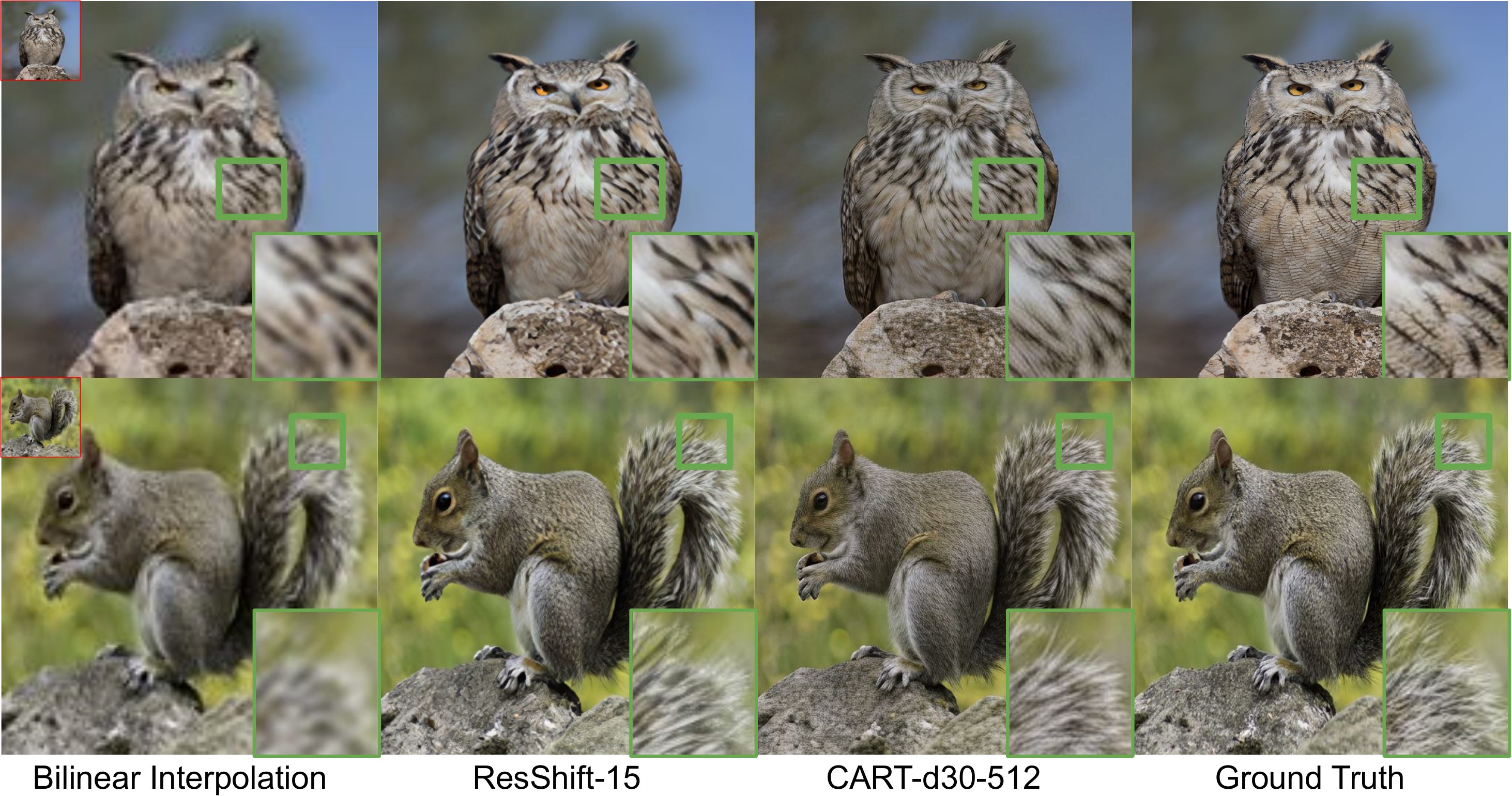}
    \captionof{figure}{Comparison of CART for super-resolution with ResShift. Zoom-in recommended to observe finer details. }
    \label{fig:sr_supp2}
\end{center}

}

}]

\begin{figure*}
\centering     
\includegraphics[width=\textwidth]{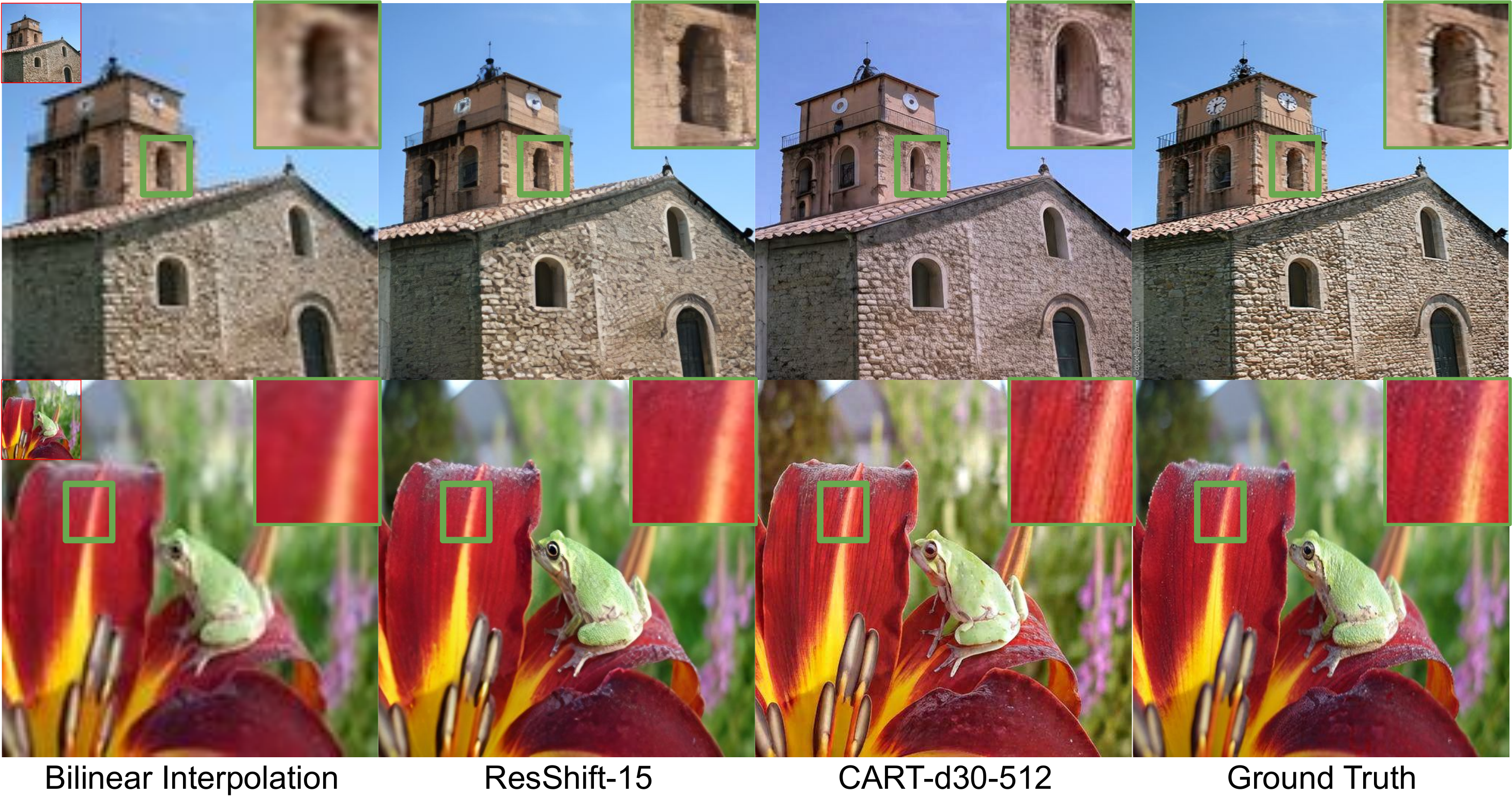}
\caption{Comparison of CART for super-resolution with ResShift. Zoom-in recommended to observe finer details.}
\end{figure*}

\begin{figure*}
\centering     
\includegraphics[width=\textwidth]{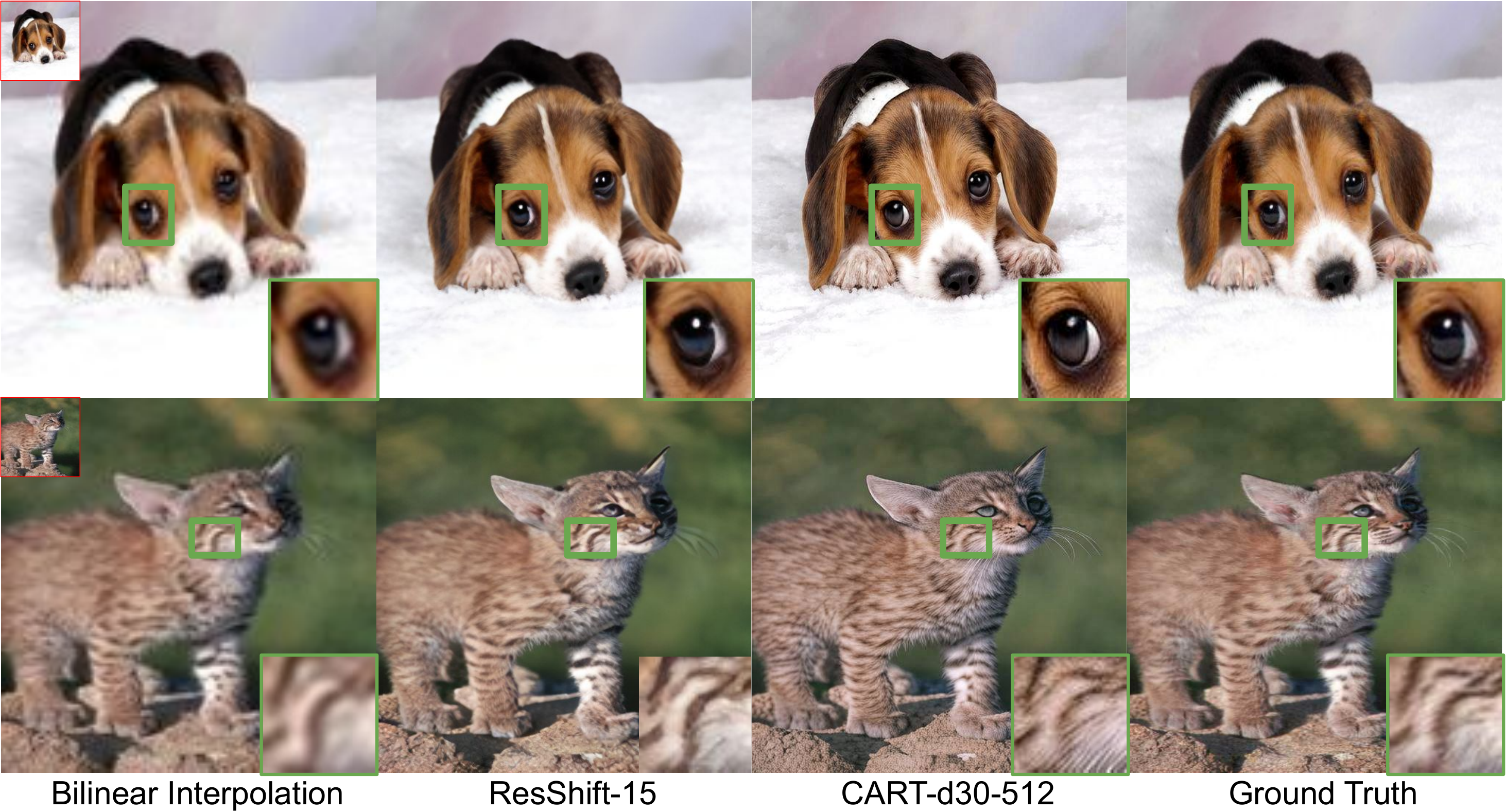}
\caption{Comparison of CART for super-resolution with ResShift. Zoom-in recommended to observe finer details.}
\end{figure*}

\begin{figure*}
\centering
    \includegraphics[width=\textwidth]{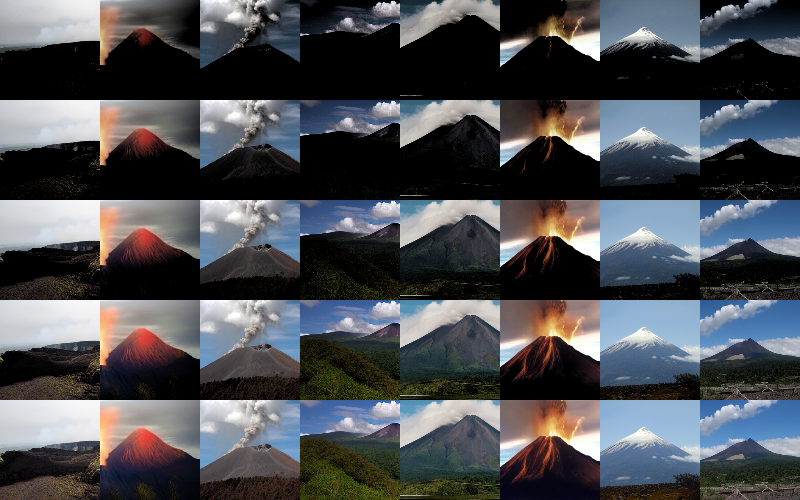}
    \caption{Generation using specularity decomposition. Top row: Base diffuse term, $2^{nd} - 5^{th}$ rows: addition of specular terms to change illumination.}
    \label{fig:specularity_decomp_2}

\end{figure*}

\twocolumn[{%
\renewcommand\twocolumn[1][]{#1}%
\justify{
\section{Specularity Decomposition for Illumination Control in Generated Images}
\label{sec:specularity}
Replacing the base-detail decomposition in Section \ref{sec:hirarchical_bd} with Specularity decomposition \cite{saini2024specularity} enables explicit control over lighting effects in the generated images. According to the dichromatic reflection model \cite{tominaga1994dichromatic} an image consists of a diffuse $\bm{A}$ and a specular $\bm{E}$: $\bm{I} = \bm{A} + \bm{E}$.  We employ a $4^{th}$-order decomposition, introducing four levels of illumination control, $\bm{I} = \bm{A}_4 + \bm{E}_4 + \bm{E}_3 + \bm{E}_2 + \bm{E}_1$. The generation process consists of 16 autoregressive steps, where the first 8 steps generate the base factor, and the subsequent 8 steps refine the image by incorporating controlled lighting variations.
Figures \ref{fig:specularity_decomp_1}-\ref{fig:specularity_decomp_2} presents examples of images generated using Specularity decomposition, demonstrating the model's ability to synthesize images with varying illumination while maintaining structural consistency.

\begin{center}
    \centering
    \captionsetup{type=figure}
    \includegraphics[width=\textwidth]{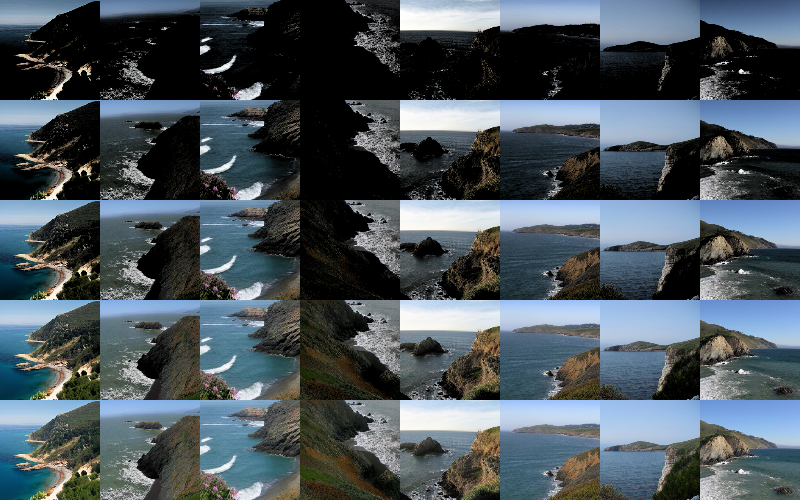}
    \captionof{figure}{Generation using specularity decomposition. Top row: Base diffuse term, $2^{nd} - 5^{th}$ rows: addition of specular terms to change illumination.}
    \label{fig:specularity_decomp_1}


\end{center}

}
}]

\begin{figure*}
\centering
    \includegraphics[width=0.99\textwidth]{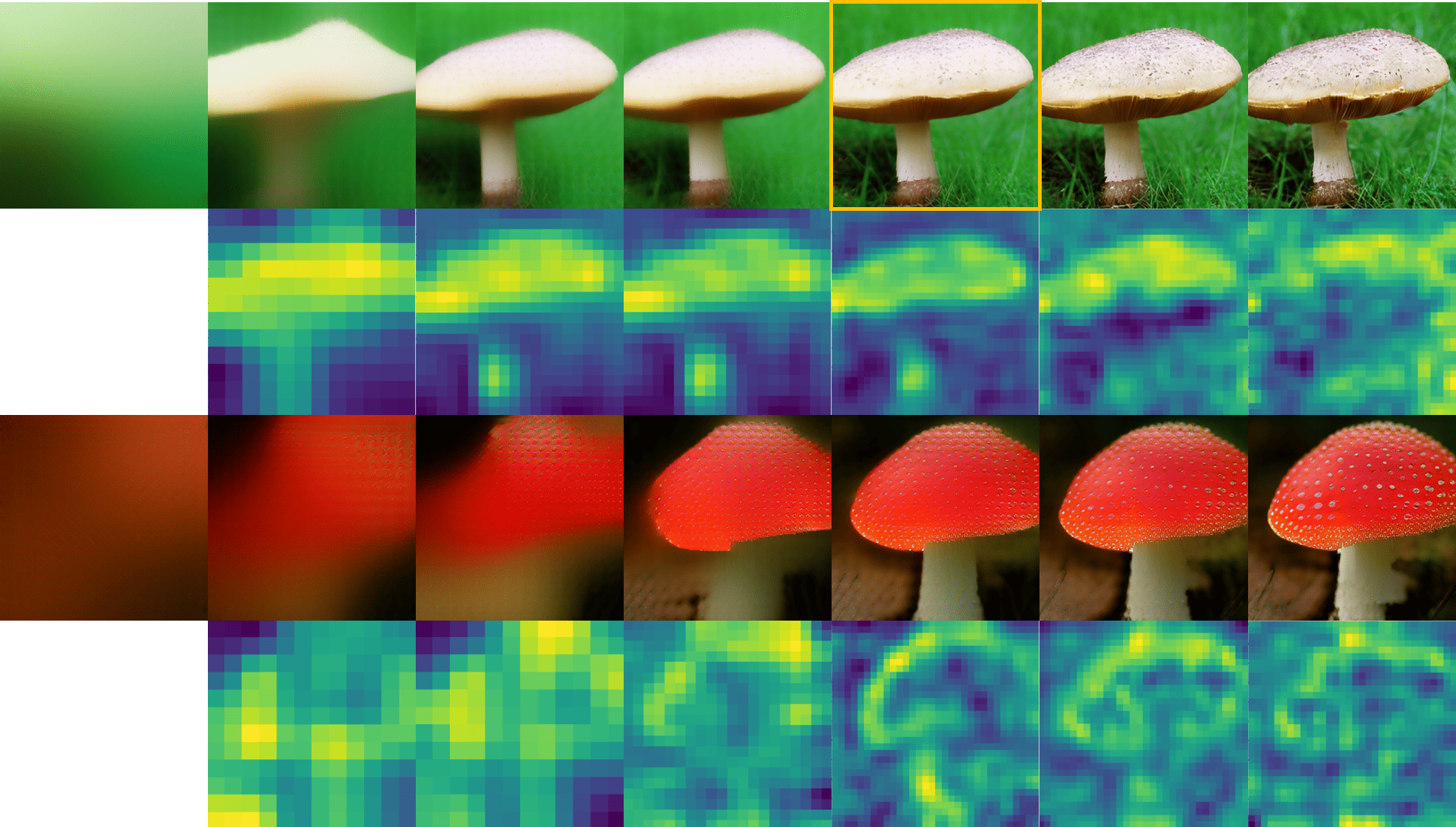}
    \caption{Top row: Intermediate visual results for CART. Base image is marked by yellow outline. $2^{nd}$ row: Self-attention maps for corresponding intermediate layers of CART. $3^{rd}$ row: Intermediate visual results for VAR. bottom row: Self-attention maps for corresponding intermediate layers of VAR.}
    \label{fig:CARTvsVAR_Results2}

\end{figure*}

\begin{figure*}
\centering
    \includegraphics[width=0.99\textwidth]{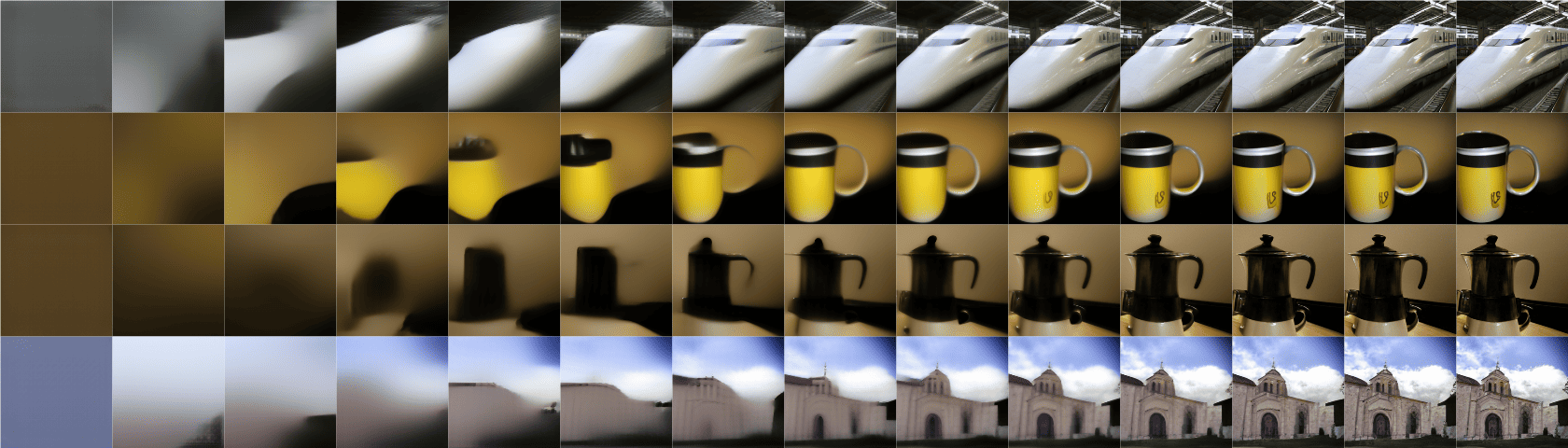}
    \caption{Stepwise generation process (left to right) of samples in CART for bullet train, coffee cup, coffee pot, and monastery classes (top to bottom)}
    \label{fig:CART_GenProc1}

\end{figure*}

\begin{figure*}
\centering
    \includegraphics[width=0.99\textwidth]{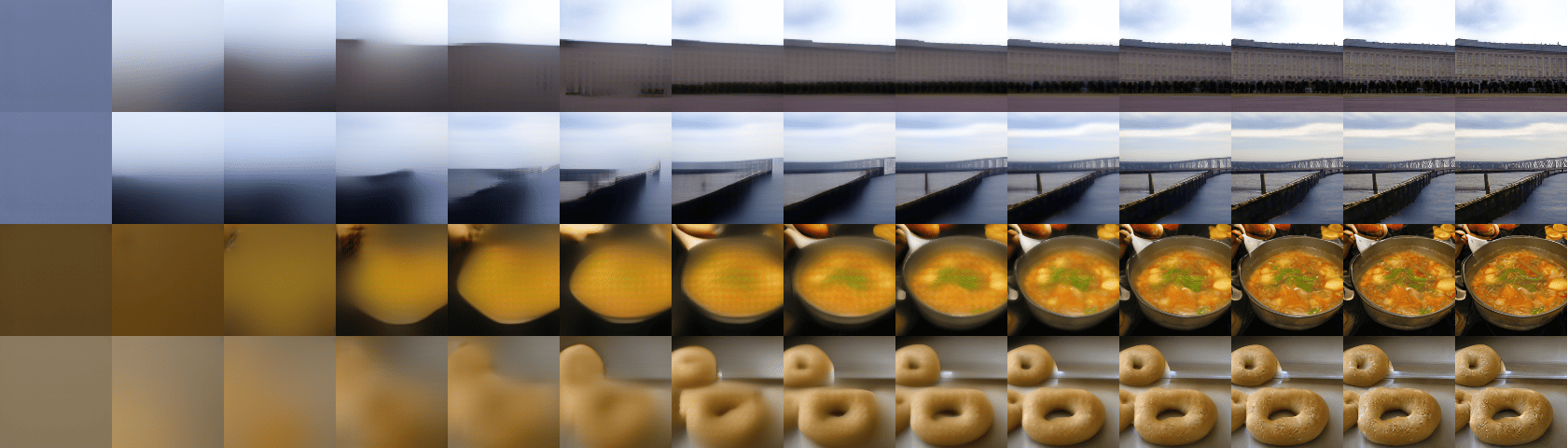}
    \caption{Stepwise generation process (left to right) of samples in CART for palace, pier, hotpot, and bagel classes (top to bottom)}
    \label{fig:CART_GenProc2}

\end{figure*}

\begin{figure*}
\centering
    \includegraphics[width=0.99\textwidth]{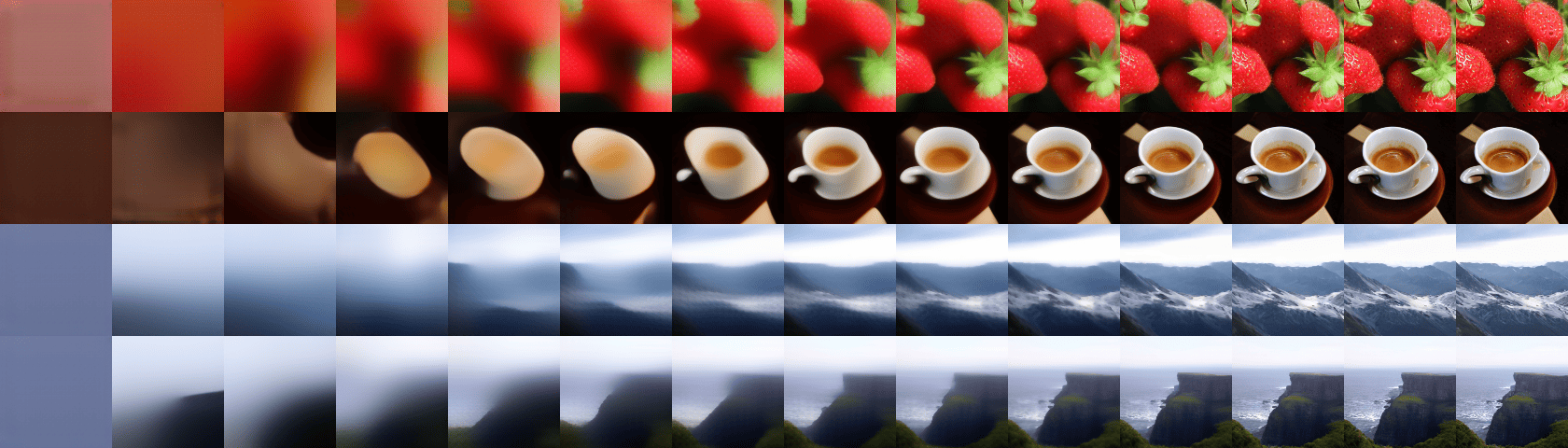}
    \caption{Stepwise generation process (left to right) of samples in CART for strawberry, espresso, alps, and cliff classes (top to bottom)}
    \label{fig:CART_GenProc3}

\end{figure*}

\begin{figure*}
\centering
    \includegraphics[width=0.99\textwidth]{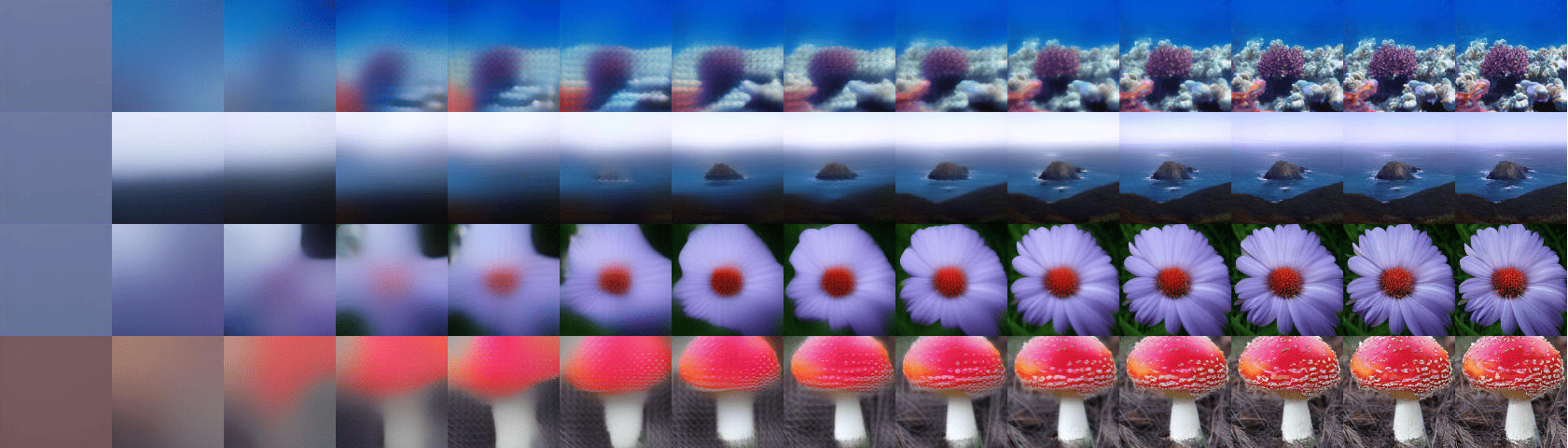}
    \caption{Stepwise generation process (left to right) of samples in CART for coral reef, foreland, daisy, and agaric classes (top to bottom)}
    \label{fig:CART_GenProc4}

\end{figure*}

\begin{figure*}
\centering
    \includegraphics[width=0.99\textwidth]{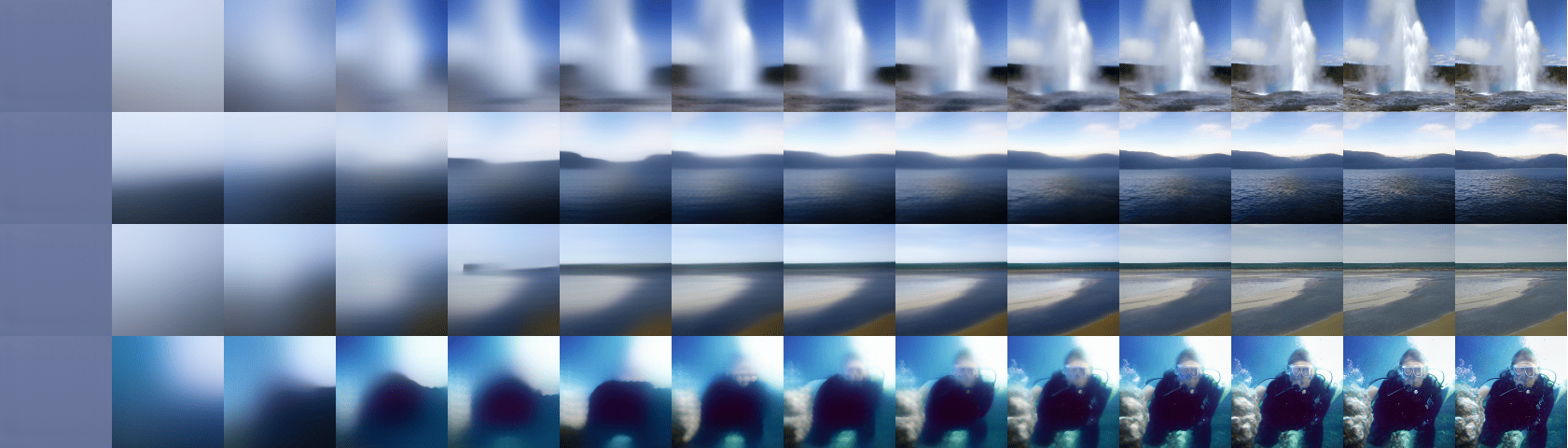}
    \caption{Stepwise generation process (left to right) of samples in CART for geyser, lake, sandbar, and scuba diver classes (top to bottom)}
    \label{fig:CART_GenProc5}

\end{figure*}

\begin{figure*}
\centering
    \includegraphics[width=0.99\textwidth]{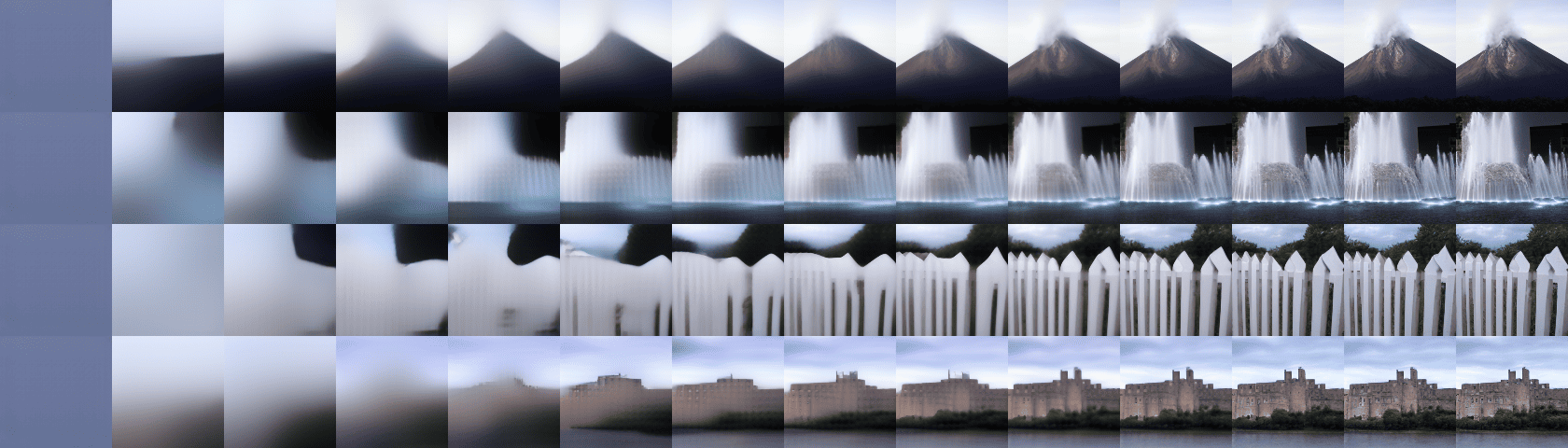}
    \caption{Stepwise generation process (left to right) of samples in CART for volcano, fountain, picket fence, and castle classes (top to bottom)}
    \label{fig:CART_GenProc6}

\end{figure*}

\twocolumn[{%
\renewcommand\twocolumn[1][]{#1}%
\justify{
\section{Generative process and comparisons with VAR}
\label{sec:generative_process}
Figures \ref{fig:CARTvsVAR_Results1}-\ref{fig:CARTvsVAR_Results2} presents a comparison of intermediate results and attention maps between CART and VAR, which utilizes multi-scale tokens. As discussed in Section \ref{sec:intro}, VAR simultaneously models both global and local features at each intermediate step, leading to an entangled representation of structure and texture. In contrast, CART first generates a smooth base image that captures only the global structure, followed by the progressive addition of fine details in later steps.

This hierarchical decomposition enables a clear separation of global and local features, facilitating high-resolution image synthesis and improving adaptability across different resolutions. Additionally, this tokenization strategy aligns more closely with human perception, where broad structures are recognized before finer details. Figures \ref{fig:CART_GenProc1}-\ref{fig:CART_GenProc6} illustrate the iterative refinement process in CART, where details are progressively introduced.

\begin{center}
    \centering
    \captionsetup{type=figure}
    \includegraphics[width=.99\textwidth]{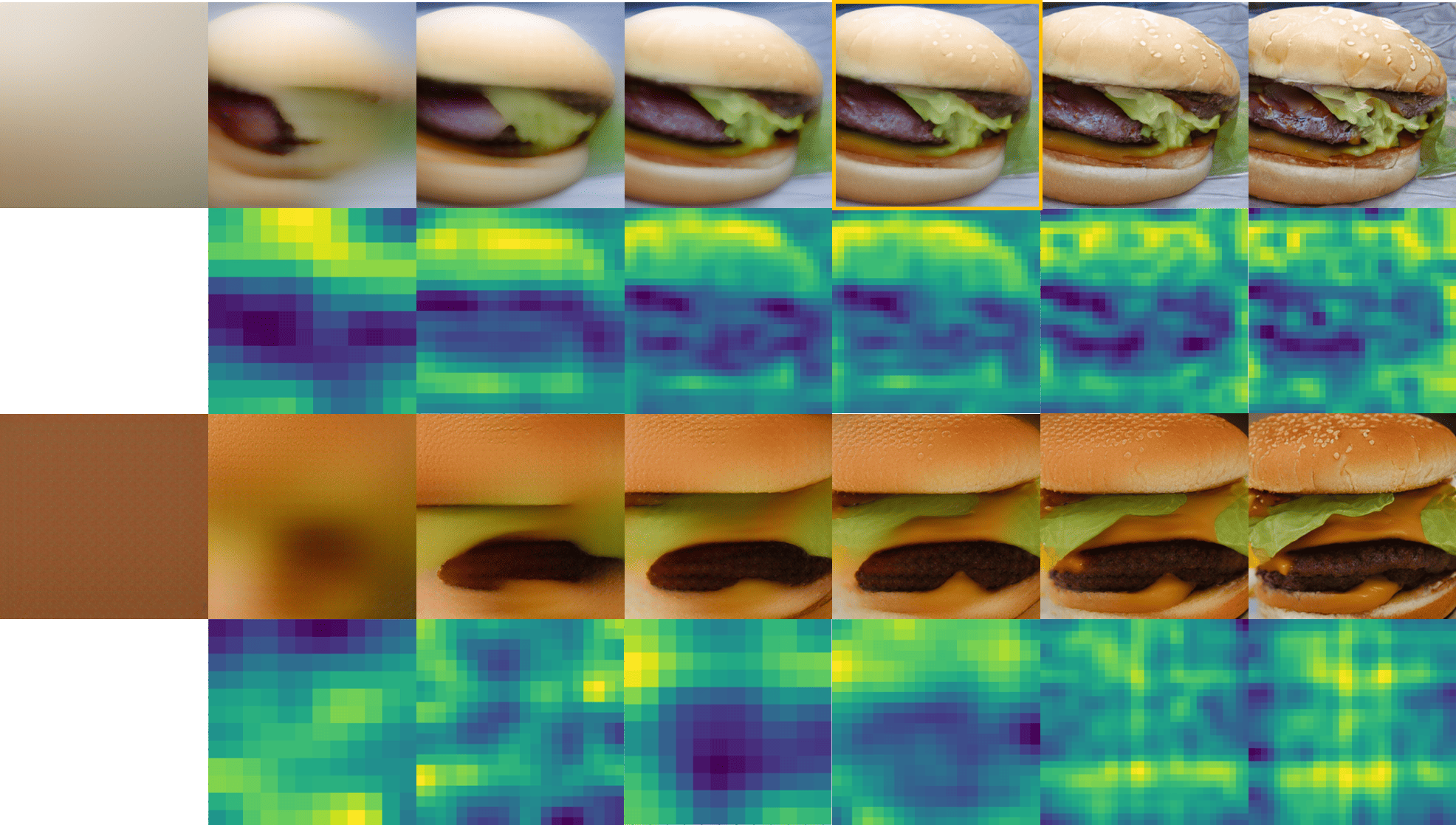}
    \captionof{figure}{Top row: Intermediate visual results for CART. Base image is marked by yellow outline. $2^{nd}$ row: Self-attention maps for corresponding intermediate layers of CART. $3^{rd}$ row: Intermediate visual results for VAR. bottom row: Self-attention maps for corresponding intermediate layers of VAR.}
    \label{fig:CARTvsVAR_Results1}

    
\end{center}%

}

}]






\newpage

\end{document}